\documentclass[10pt,twocolumn,letterpaper]{article}

\usepackage[pagenumbers]{cvpr}      

\usepackage[dvipsnames]{xcolor}

\definecolor{cvprblue}{rgb}{0.21,0.49,0.74}
\usepackage[pagebackref,breaklinks,colorlinks,citecolor=cvprblue]{hyperref}

\usepackage{float}
\usepackage{multirow}

\usepackage[marginal]{footmisc}

\def\paperID{XXXX} 
\def\confName{CVPR}
\def\confYear{2024}

\title{
    Sketch2NeRF: Multi-view Sketch-guided Text-to-3D Generation
}

\author{
Minglin Chen$^{1}$\footnotemark[1]
\qquad
Weihao Yuan$^{2}$
\qquad
Yukun Wang$^{1}$
\qquad
Zhe Sheng$^{2}$ \\
\qquad
Yisheng He$^{2}$
\qquad
Zilong Dong$^{2}$
\qquad
Liefeng Bo$^{2}$
\qquad
Yulan Guo$^{1}$\footnotemark[2]
\vspace{0.1cm}\\
{\normalsize $^{1}$ The Shenzhen Campus of Sun Yat-sen University, Sun Yat-sen University \quad $^{2}$ Alibaba Group}
}

\begin{document}

\twocolumn[{
\maketitle
\vspace{-26pt}
\begin{figure}[H]
    \hsize=\textwidth
    \centering
    \includegraphics[width=2.0\linewidth]{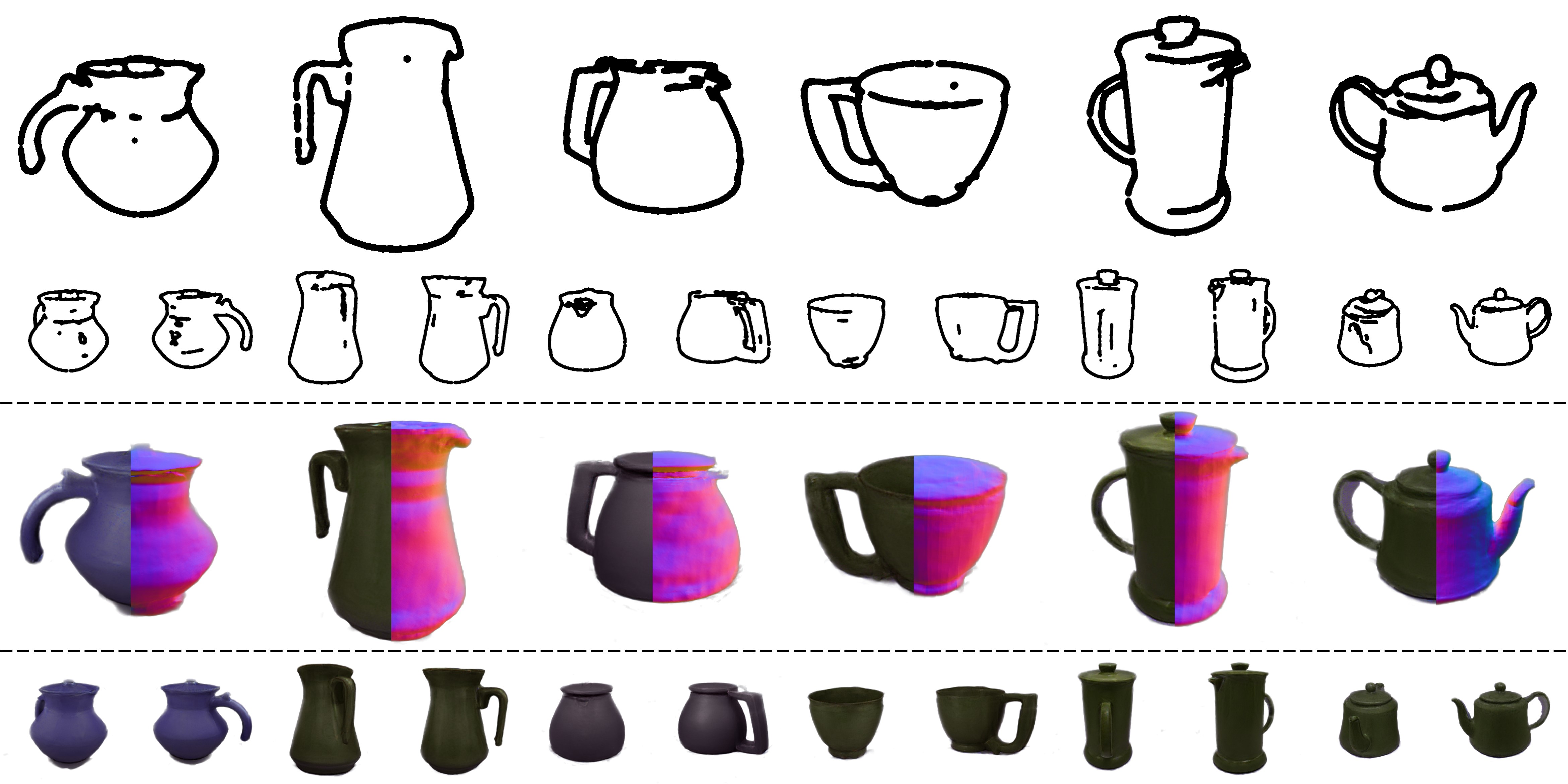}
    \caption{
        \textbf{Sketch2NeRF} is a sketch-guided text-to-3D generative model that produces high-fidelity 3D objects resembling multi-view sketches.
        \textit{Top}: our method can use an arbitrary number of sketches (usually more than 3) as input.
        \textit{Middle}: generated 3D objects (shown as rendered RGB and normal images) of which the shapes are controlled by input sketches.
        \textit{Bottom}: rendered RGB images at novel views.
        Note that, these 3D objects are generated using the same prompt of \textit{“a teapot”}.
    }
    \label{fig:tesser}
\end{figure}
}]

\footnotetext[1]{Work done during an internship supervised by Weihao Yuan at Alibaba Group.}
\footnotetext[2]{Corresponding author: \textcolor{magenta}{guoyulan@mail.sysu.edu.cn}}

\begin{abstract}
Recently, text-to-3D approaches have achieved high-fidelity 3D content generation using text description.
However, the generated objects are stochastic and lack fine-grained control.
Sketches provide a cheap approach to introduce such fine-grained control. Nevertheless, it is challenging to achieve flexible control from these sketches due to their abstraction and ambiguity. 
In this paper, we present a multi-view sketch-guided text-to-3D generation framework (namely, Sketch2NeRF) to add sketch control to 3D generation.
Specifically, our method leverages pretrained 2D diffusion models (e.g., Stable Diffusion and ControlNet) to supervise the optimization of a 3D scene represented by a neural radiance field (NeRF).
We propose a novel synchronized generation and reconstruction method to effectively optimize the NeRF.
In the experiments, we collected two kinds of multi-view sketch datasets to evaluate the proposed method.
We demonstrate that our method can synthesize 3D consistent contents with fine-grained sketch control while being high-fidelity to text prompts.
Extensive results show that our method achieves state-of-the-art performance in terms of sketch similarity and text alignment.
\end{abstract}
\section{Introduction}

Distilling pretrained large-scale text-to-image diffusion models~\cite{rombach2022high,saharia2022photorealistic} to neural radiance fields (NeRF)~\cite{mildenhall2021nerf,barron2022mip,muller2022instant,shen2021deep,chen2022tensorf,fridovich2022plenoxels,sun2022direct} has emerged as a powerful paradigm for text-conditioned 3D object generation~\cite{wang2023prolificdreamer,poole2022dreamfusion,lin2023magic3d,wang2023score,metzer2022latent,chen2023fantasia3d,tsalicoglou2023textmesh}.
The success of these models comes from the development of score distillation sampling (SDS)~\cite{poole2022dreamfusion}, which is further improved within a variational framework like variational score distillation (VSD)~\cite{wang2023prolificdreamer}.
In spite of the high-fidelity, diverse, and 3D-consistent results achieved by text-to-3D methods, existing works lack fine-grained controllable ability during the generation procedure.

To achieve fine-grained controllable 3D generation, a straightforward way is prompt-based approaches that construct a fine-grained prompt with more specific words. 
Existing text-to-3D approaches~\cite{jain2022zero,lin2023magic3d} have shown the potential controllability of compositional prompts that combine two concrete concepts in a prompt template for 3D generation.
Still, prompt-based approaches are difficult to achieve fine-grained control since a certain spatial geometry and concept is difficult to describe in words.
In addition, several shape-guided approaches~\cite{metzer2022latent,cao2023dreamavatar,chen2023fantasia3d} use a coarse 3D model (e.g, SMPL model~\cite{SMPL:2015}, morphable face model~\cite{li2020learning}, and simple 3D primitives) as initial geometry to guide the generation.
The generated model is highly sensitive to the initial geometry, while the fine geometry model is not easy to obtain.
More recently, image-based approaches explore to generate a 3D asset using one image~\cite{liu2023zero,tang2023make} or a few images~\cite{raj2023dreambooth3d}.
However, it is challenging for these methods to achieve fine-grained control in the 3D space with only modifications in the images.
Because humans find it hard to draw a real image with rich texture.
Therefore, fine-grained control in 3D generation which guides spatial properties of the 3D asset remains an open problem.

In 2D image generation, sketches offer a feasible and promising approach to provide controllable structure cues,
but it is challenging to generate plausible 3D assets from the sketch-based constraint due to its simplicity and abstraction.
Some pioneering methods make attempts to generate 3D shapes from single-view sketches~\cite{sanghi2023sketch}. Nevertheless, the geometry constraints provided by the one-view sketch are insufficient to synthesize plausible 3D objects.

In this paper, we make the first attempt to tackle the multi-view sketch-guided 3D object generation problem by proposing a novel framework (namely Sketch2NeRF).
We first employ a neural radiance field (NeRF)~\cite {mildenhall2021nerf} to represent the underlying 3D object,
and then leverage the pretrained 2D diffusion models (i.e., Stable Diffusion and ControlNet) to supervise the optimization of NeRF.
In particular, we propose a novel synchronized generation and reconstruction mechanism to effectively optimize the NeRF. 
Furthermore, the annealed time schedule is introduced to improve the quality of the generated object.
For performance evaluation of the proposed sketch-guided 3D generation framework, we introduce sketch-based datasets and metrics to evaluate the controllability of 3D object generation. 
Experimental results demonstrate that our method can fine-grainedly control the 3D generation based on sketches, which offers flexible controllability to existing text-to-3D methods.

To the best of our knowledge, this is the first work for controllable generation of 3D shapes from multi-view sketches.
Our main contributions in this paper are summarized as follows:
\begin{itemize}
\setlength{\itemsep}{0pt}
    \item We propose a novel framework for multi-view sketch-guided 3D object generation, which enables fine-grained control during generation.
    \item We leverage sketch-conditional 2D diffusion models to guide the 3D generation, which eliminates the need for a large sketch-3D paired dataset.
    \item We collect sketch-based generation datasets and evaluation metrics to show the superior fine-grained control ability of the proposed method over existing text-to-3D counterparts.
\end{itemize}
\section{Related Work}
In this section, we first review the 3D generation approaches and controllable generation. Then, we briefly describe the development of sketch-based generation.

\subsection{3D Generation}

\noindent\textbf{Prompt-based}
With the emergence of pretrained text-conditioned image generative diffusion models~\cite{saharia2022photorealistic}, lifting these 2D models to achieve text-to-3D generation has gained popularity.
Poole~\textit{et al}.~\cite{poole2022dreamfusion} proposed DreamFusion with SDS for text-to-3D generation. 
The lack of 3D awareness in the 2D diffusion models destabilizes score distillation-based methods from reconstructing a plausible 3D object.
To address this issue, Seo~\textit{et al}.~\cite{seo2023let} proposed 3DFuse, a novel framework that incorporates 3D awareness into pretrained 2D diffusion models, enhancing the robustness and 3D consistency of score distillation-based methods.
Chen~\textit{et al}.~\cite{chen2023fantasia3d} proposed Fantasia3D for high-quality text-to-3D content creation.
This method disentangles modeling and learning of geometry and appearance to achieve high-resolution supervision from 2D diffusion models.
Xu~\textit{et al}.~\cite{xu2023dream3d} proposed a simple yet effective approach that directly bridges the text and image modalities with a powerful text-to-image diffusion model.
Wang~\textit{et al}.~\cite{wang2023prolificdreamer} proposed to model the 3D parameter as a random variable instead of a constant as in SDS and present VSD to explain and address over-saturation, over-smoothing, and low-diversity problems.

\noindent\textbf{Image-based}
3D generation from a single image or a few images is an ill-posed problem.
Previously, PixelNeRF~\cite{yu2021pixelnerf} and GRF~\cite{Trevithick_2021_ICCV} are proposed to employ pretrained visual models (e.g., CNN~\cite{He_2016_CVPR}) as a prior.
Recently, one-shot or few-shot 3D generation approaches explore diffusion models as a prior.
Deng~\textit{el al}.~\cite{deng2023nerdi} proposed NeRDi, a single-view NeRF generation framework with general image priors from 2D diffusion models.
As an alternative, Wimbauer~\textit{et al}.~\cite{wimbauer2023behind} proposed to predict an implicit density field from a single image.
Gu~\textit{et al}.~\cite{gu2023nerfdiff} proposed NeRFDiff, which addresses the occlusion issue by distilling the knowledge of a 3D-aware conditional diffusion model (CDM) into NeRF through synthesizing and refining a set of virtual views at test-time.
Liu~\textit{et al}.~\cite{liu2023zero} proposed Zero-1-to-3 to generate a 3D object from a single image by learning to control the camera perspective in large-scale diffusion models.
Tang~\textit{et al}.~\cite{tang2023make} proposed Make-it-3D to create high-fidelity 3D content from a single image using pretrained diffusion priors.

\subsection{Controllable Generation}
\noindent\textbf{2D Image}
The 2D generative approaches~\cite{metzer2022latent} based on the diffusion probabilistic model~\cite{ho2020denoising} can synthesize high-fidelity images according to text description.
To append more controllable conditions (e.g., Canny edge, human pose, semantic map, depth) in generation, 
ControlNet~\cite{zhang2023adding}, T2I-Adapter~\cite{mou2023t2i}, and Uni-ControlNet~\cite{zhao2023uni} are proposed to learn task-specific conditions.
Besides, DreamBooth~\cite{ruiz2023dreambooth} controls the content of synthesized images using a few images of the same subject.
Recently, DragDiffusion~\cite{shi2023dragdiffusion} is proposed to achieve point-based interactive image generation.

\noindent\textbf{3D Shape}
These controllable 3D generation approaches based on 2D generative diffusion models use either coarse intialized geometry or controllable 2D models.
Latent-NeRF~\cite{metzer2022latent} and Fantasia3D~\cite{chen2023fantasia3d} use a coarse 3D model (e.g., simple 3D primitives) as initialized geometry, which is further refined and textured during generation.
In the other direcitons, DreamBooth3D~\cite{raj2023dreambooth3d} leverages DreamBooth~\cite{ruiz2023dreambooth} to specify the content of generated a 3D asset using 3$\sim$5 images, while Instruct-NeRF2NeRF~\cite{haque2023instruct} and Instruct 3D-to-3D~\cite{kamata2023instruct} employs Instruct-Pix2Pix~\cite{brooks2023instructpix2pix} to edit a pretrianed NeRF.

\begin{figure*}[hbpt]
    \centering
    \includegraphics[width=\linewidth]{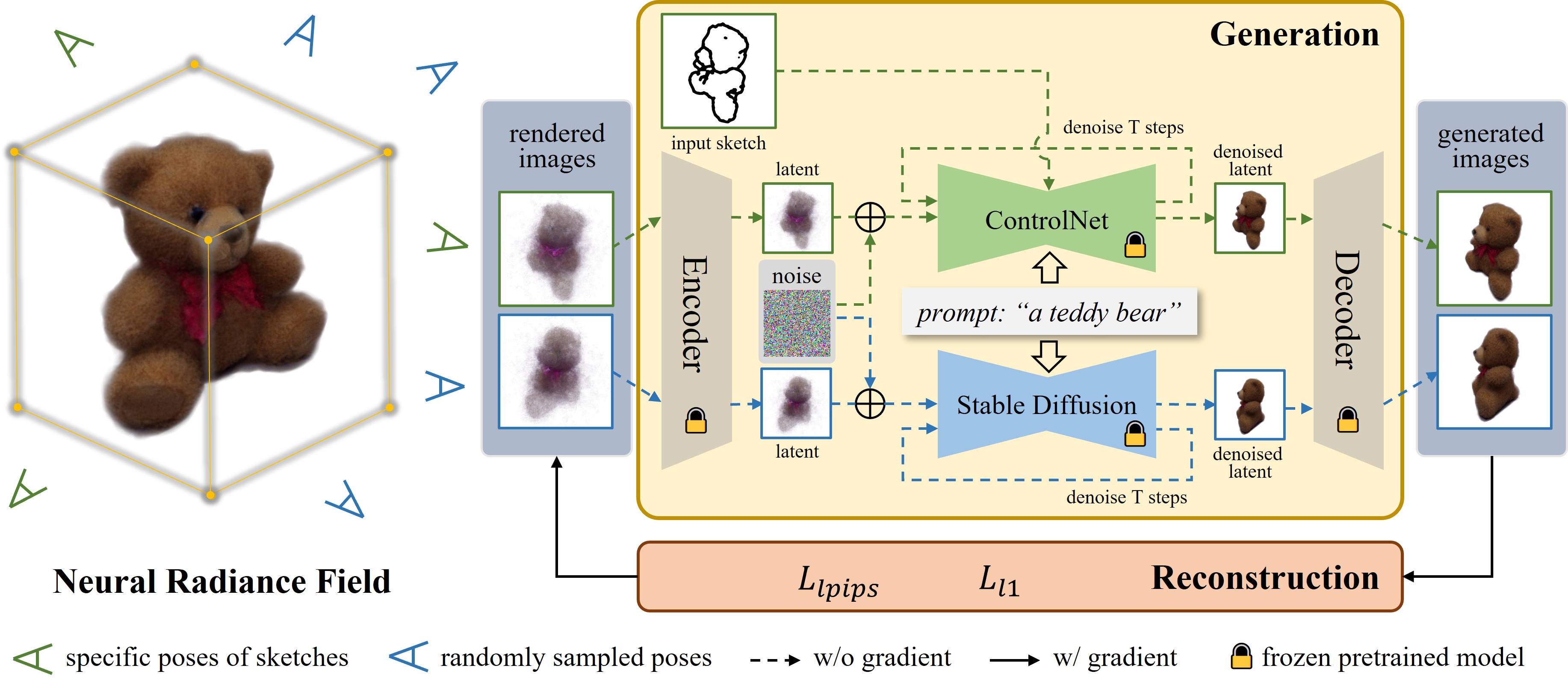}
    \caption{
       \textbf{Sketch2NeRF Overview}. 
       We represent a 3D object using a neural radiance field (NeRF) which is optimized based on the proposed synchronized generation and reconstruction optimization method.
       At the generation stage, the ControlNet is used to generate real images at specific poses of sketches, while the Stable Diffusion is employed to generate real images at randomly sampled poses.
       At the reconstruction stage, we update the NeRF parameters such that the reconstruction loss between the generated and rendered images is minimized.
    }
    \label{fig-framework}
    \vspace{-0.4cm}
\end{figure*}

\subsection{Sketch-based 3D Generation}
3D generation from the sketch is a challenging task, due to the high-level abstraction and ambiguity of sketches.
He~\textit{et al}.~\cite{he2023sketch2cloth} proposed Sketch2Cloth, a sketch-based 3D garment generation system using the unsigned distance fields from the user's sketch input.
Applying sketches to neural radiance fields~\cite{mildenhall2021nerf} is also challenging due to the inherent uncertainty for 3D generation with coarse 2D constraints, a significant gap in content richness, and potential inconsistencies for sequential multi-view editing given only 2D sketch inputs.
To address these challenges, Gao~\textit{et al}.~\cite{lin2023sketchfacenerf} present SketchFaceNeRF, a novel sketch-based 3D facial NeRF generation and editing method, to produce free-view photo-realistic images.
Recently, Sanghi~\textit{et al}.~\cite{sanghi2023sketch} proposed Sketch-A-Shape to generate a 3D shape from a single sketch by leveraging the pretrained CLIP model~\cite{radford2021learning}.
Wu~\textit{et al}.~\cite{wu2023sketch} proposed a sketch-guided diffusion model for colored point cloud generation.
Mikaeili~\textit{et al}.~\cite{mikaeili2023sked} present SKED to edit a NeRF using the sketch cues.

Existing sketch-based 3D generation methods are inference-based and have the following limitations: 1) can only input a single sketch; 2) require a large amount of data to train the network; and 3) the generated objects are limited to a few specific categories (e.g., face~\cite{lin2023sketchfacenerf}).
By leveraging the pretained 2D diffusion models, our method is an optimization-based 3D generation method that achieves open-vocabulary object generation from an arbitrary number of sketches as input without any training data.
\section{Methodology}
Given a set of $N$ multiview sketch images $\{S_n\}_1^N$ with poses $\{\pi_n\}_1^N$, and a text prompt $T$, our goal is to generate a 3D asset that resembles specific sketches at all given angles.
We employ an optimization-based framework that leverages pretrained 2D diffusion models (Sec.~\ref{Guidance}) as guidance to fit a NeRF-based 3D representation (Sec.~\ref{3DRepresentation}).
The optimization details are described in Sec.~\ref{Optimization}.
The overview of our method is shown in Fig.~\ref{fig-framework}.

\subsection{3D Representation}
\label{3DRepresentation}
We represent a 3D object using the neural radiance fields (NeRF)~\cite{mildenhall2021nerf}, which is flexible and capable of rendering photo-realistic images~\cite{barron2021mip,barron2022mip,verbin2022ref,barron2023zip}.
Suppose the 3D object is located at the center of a bounded region (e.g., $[-1,1]^3$).
For each 3D point $\pmb{x} = (x,y,z)$ and its view direction $\pmb{d} = (\theta, \phi)$, NeRF represents its density and view-dependent color as a continuous function:
\begin{equation}
\label{NeRF}
    (\sigma, \pmb{c}) = f_\theta(\pmb{x}, \pmb{d}),
\end{equation}
where $\theta$ is the parameters of the continuous function $f$.
For faster coverage and efficient memory cost, we implement $f$ following~\cite{muller2022instant}.

We employ the volume rendering technique~\cite{mildenhall2021nerf} to render the image with pose.
In particular, the ray $\pmb{r} = \pmb{o} + t \cdot \pmb{d}$ is calculated according to the image pose for each pixel.
Then, we sample points along the ray as $\{\pmb{x}_i\}$.
The density $\sigma_i$ and color $\pmb{c}_i$ of the sampled points $\pmb{x}_i$ are evaluated using Eq. (\ref{NeRF}).
Finally, we obtain the ray color (i.e., the pixel color) by ray marching:
\begin{equation}
    C(\pmb{r})=\sum_i T_i (1 - \rm{exp}(- \sigma_i \delta_i)) \pmb{c}_i,
\end{equation}
where $T_i = \rm{exp}(- \sum_{j=1}^{i-1} \sigma_j \delta_j)$, and $\delta_i = t_{i+1}-t_i$ is the interval between adjacent sampled points.
Image rendering is performed by repeating the rendering of each pixel.

\subsection{Sketch-conditioned Guidance}
\label{Guidance}
Given a randomly initialized NeRF, we iteratively update its weights using a 2D pretrained diffusion model as the guidance.
To incorporate the constraint of multi-view sketches, we employ a pretrained 2D sketch-conditioned diffusion model.

In particular, the sketch-conditioned diffusion model comprises of a variational auto-encoder (VAE) and a denoising network.
The VAE consists of an encoder $\mathcal{E}$ and a decoder $\mathcal{D}$, where the encoder transforms an image $\pmb{x}$ to a latent variable as $\pmb{z} = \mathcal{E}(\pmb{x})$ and the decoder reconstruct the image from the latent variable as $\hat{\pmb{x}} = \mathcal{D}(\pmb{z})$.
The denoising network implemented by an improved U-Net~\cite{ronneberger2015u} is used to generate the latent variable $\pmb{z}_0$ of images by iteratively denoising from a Gaussian noise $\pmb{z}_T$.
The denoising network estimates the noise $\hat{\pmb{\epsilon}}_{t}$ at level $t$: 
\begin{equation}
    \hat{\pmb{\epsilon}}_t := \epsilon_{\phi}(\pmb{z}_t;t,\pmb{y}),
\end{equation}
where $\pmb{y}$ is the conditional prompt.
We improve the generation quality with classifier-free guidance~\cite{ho2022classifier}, thus the estimated noise is obtained as:
\begin{equation}
    \hat{\pmb{\epsilon}}_t := (1 + \omega ) \cdot \epsilon_{\phi}(\pmb{z}_t;t,\pmb{y}) - \omega \cdot \epsilon_{\phi}(\pmb{z}_t;t,\emptyset),
\end{equation}
where $\omega$ ($\omega = 7.5$ in our experiments) is the weight for classifier-free guidance, and $\emptyset$ denotes the empty prompt.

We incorporate the sketch condition in generation following~\cite{zhang2023adding}.
The sketch condition is a binary image $\pmb{S} \in \{0,1\}^{H \times W}$, where $0$ denotes the canvas and $1$ denotes the stroke of the sketch.
The latent denoising network has an additional sketch condition:
\begin{equation}
    \hat{\pmb{\epsilon}}_t := \epsilon_{\phi}(\pmb{z}_t;t,\pmb{y},\pmb{S}).
\end{equation}

\subsection{Optimization}
\label{Optimization}
A naive approach for the 3D generation with multi-view sketch constraints is replacing Stable Diffusion in the existing text-to-3D methods~\cite{wang2023prolificdreamer} with ControlNet.
In practice, we found that such methods lead to severely degraded performance (as shown in Fig.~\ref{fig-results-qualitative-comparisons}).
This is because the score-based distillation optimization method~\cite{poole2022dreamfusion,wang2023prolificdreamer} is prone to focusing on edges rather than the overall content with the Sketch-conditioned ControlNet guidance.
To effectively optimize NeRF with ControlNet guidance, we propose a synchronized generation and reconstruction method.

\noindent{\textbf{Generation.}}
We leverage pretrained ControlNet and Stable Diffusion to generate real images at the specific poses of sketches and randomly sampled poses, respectively.

1) Specific poses of sketches.
For specific poses $\pi_s$ with the sketch image constraint, we render an image $\pmb{I}_s$ from the NeRF.
The latent variable $\pmb{z}$ of the image is added to the Gaussian noise at level $t$:
\begin{equation}
    \label{eq.noisy_latent}
    \pmb{z}_\text{noise} = \sqrt{\bar\alpha_t} \pmb{z} + \sqrt{1-\bar\alpha_t} \pmb{\epsilon},
\end{equation}
where $t\sim\mathcal{U}(0,1)$, $\pmb\epsilon$ is the Gaussian noise, and $\bar\alpha_t := \prod_{s=1}^{t}\alpha_t$, $\alpha_s$ is a variance schedule for adding Gaussian noise to the data.

Given the noisy latent variable $\pmb{z}_\text{noise}$, we employ the sketch-conditioned ControlNet to iteratively denoise the latent variable for $T$ steps using DDIM~\cite{song2020denoising}, resulting in a latent variable $\pmb{z}_0$.
The generated image is obtained by decoding the latent variable as $\hat{\pmb{I}}_s$.

\begin{figure}[t]
    \centering
    \begin{subfigure}{0.48\linewidth}
        \begin{minipage}[t]{\linewidth}
            \includegraphics[width=\linewidth]{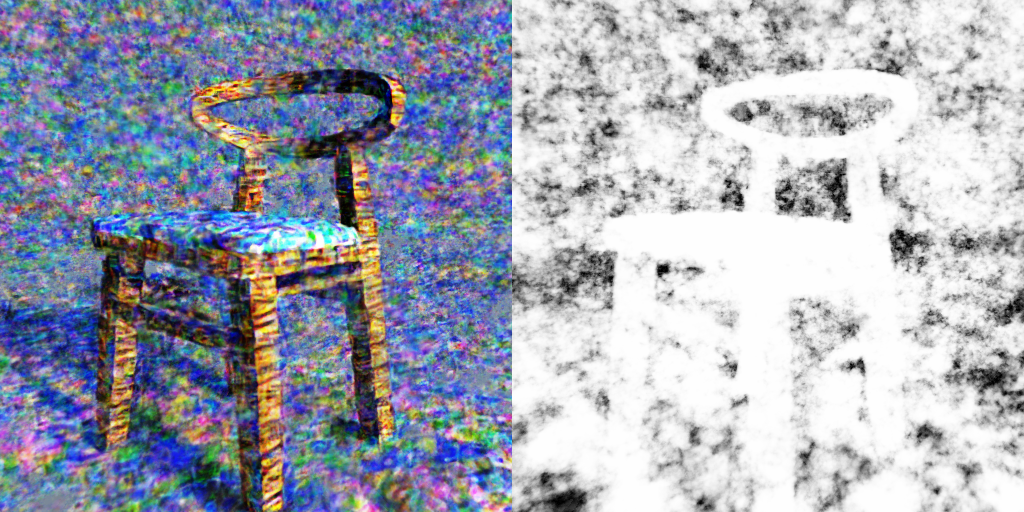}
        \end{minipage}
        \caption{w/o}
    \end{subfigure}
    \begin{subfigure}{0.48\linewidth}
        \begin{minipage}[t]{\linewidth}
            \includegraphics[width=\linewidth]{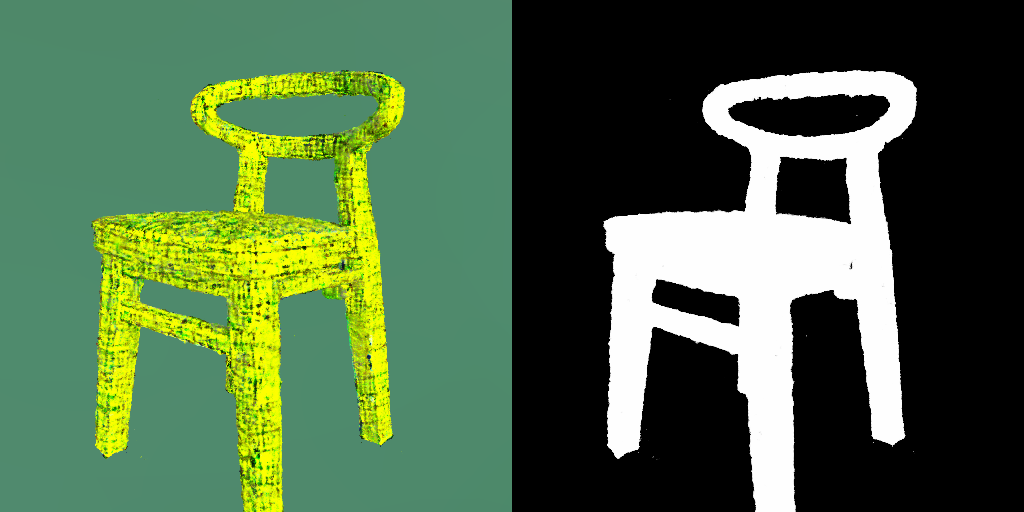}
        \end{minipage}
        \caption{w/}
    \end{subfigure}
    \caption{
        The rendered RGB and opacity images of the generated 3D objects w/o and w/ the random viewpoint regularization.
        The random viewpoint regularization effectively eliminates the \textit{near-plane} artifacts and the \textit{floaters} for the generated 3D objects. 
    }
    \label{fig-method-random-viewpoints}
    \vspace{-0.4cm}
\end{figure}

2) Randomly sampled poses.
The number of sketch images is usually small, and insufficient to reconstruct a NeRF.
Training NeRF with only supervision from sketch images produces \textit{near-plane} artifacts and many \textit{floaters} (Fig.~\ref{fig-method-random-viewpoints} (a)).
We introduce random viewpoint regularization to eliminate this issue (Fig.~\ref{fig-method-random-viewpoints} (b)).
Specifically, we randomly sample the pose $\pi_r$ at the upper hemisphere.
The image $\pmb{I}_r$ is rendered from the NeRF.
The corresponding generated image $\hat{\pmb{I}}_r$ is obtained using a similar way as in the specific poses of sketches, except for using Stable Diffusion.

\begin{figure}[t]
    \centering
    \begin{subfigure}{0.32\linewidth}
        \begin{minipage}[t]{\linewidth}
            \includegraphics[width=\linewidth]{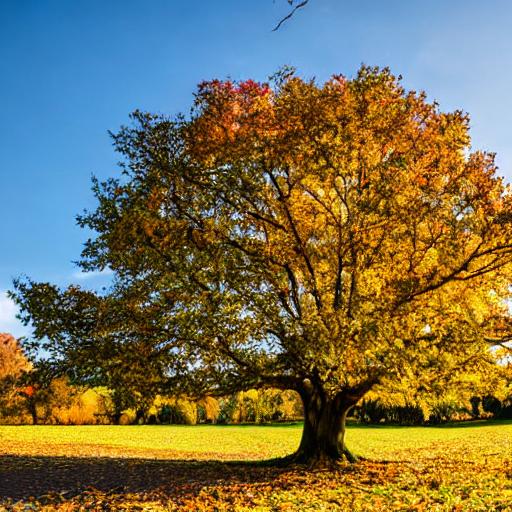}
        \end{minipage}
        \caption*{Original}
    \end{subfigure}
    \begin{subfigure}{0.32\linewidth}
        \begin{minipage}[t]{\linewidth}
            \includegraphics[width=\linewidth]{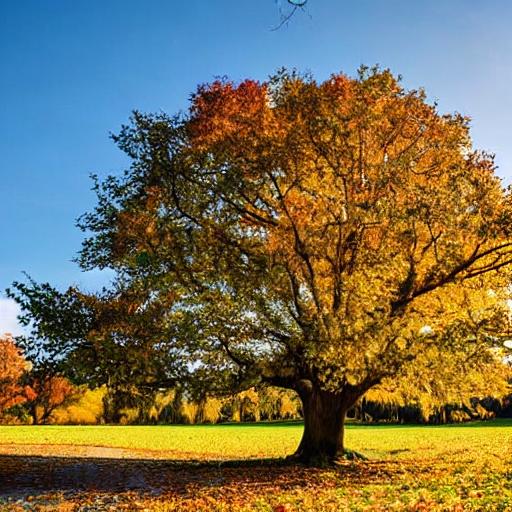}
        \end{minipage}
        \caption*{$t = 0.2$}
    \end{subfigure}
    \begin{subfigure}{0.32\linewidth}
        \begin{minipage}[t]{\linewidth}
            \includegraphics[width=\linewidth]{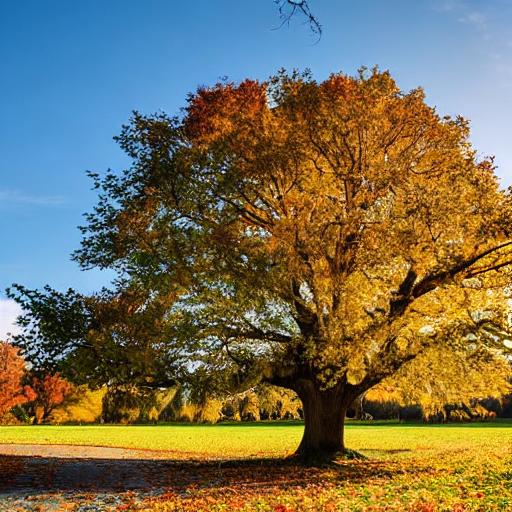}
        \end{minipage}
        \caption*{$t = 0.4$}
    \end{subfigure}
    \begin{subfigure}{0.32\linewidth}
        \begin{minipage}[t]{\linewidth}
            \includegraphics[width=\linewidth]{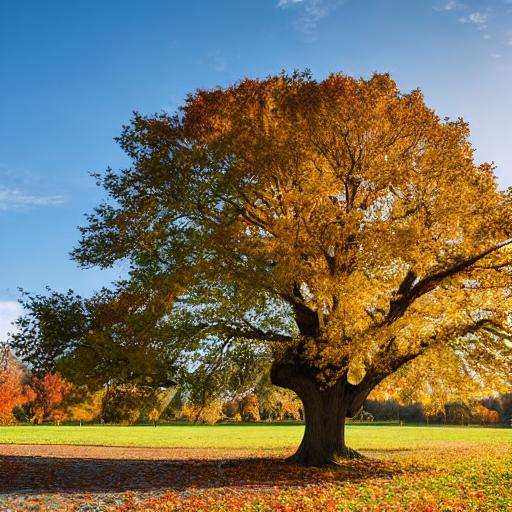}
        \end{minipage}
        \caption*{$t = 0.6$}
    \end{subfigure}
    \begin{subfigure}{0.32\linewidth}
        \begin{minipage}[t]{\linewidth}
            \includegraphics[width=\linewidth]{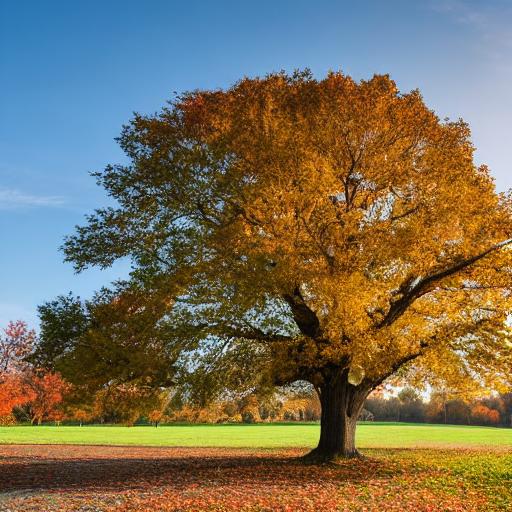}
        \end{minipage}
        \caption*{$t = 0.8$}
    \end{subfigure}
    \begin{subfigure}{0.32\linewidth}
        \begin{minipage}[t]{\linewidth}
            \includegraphics[width=\linewidth]{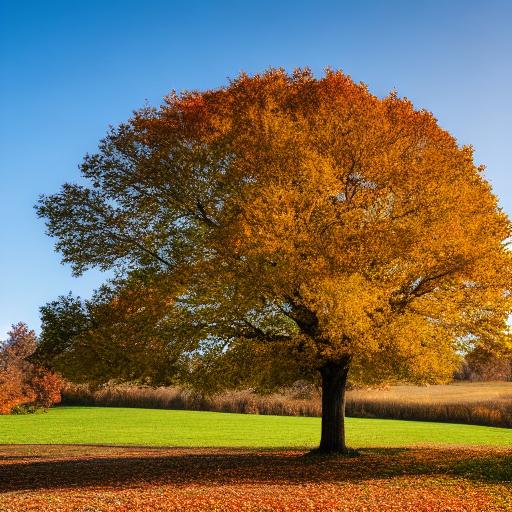}
        \end{minipage}
        \caption*{$t = 0.98$}
    \end{subfigure}
    \caption{
         Images generated with different levels of noise.
         The generated images are far different from the original when the added noise is large (e.g., $t=0.98$).
    }
    \label{fig-method-annealed-time-schedule}
    \vspace{-0.4cm}
\end{figure}

3) Annealed time schedule.
During optimization, previous text-to-3D approaches~\cite{poole2022dreamfusion} add noise to the NeRF-rendered image at uniformly distributed noise level $t\sim \mathcal{U}(t_\text{min},t_\text{max})$, where $t_\text{min}$ and $t_\text{max}$ are close to 0 and 1, respectively.
However, the generation cannot converge at the last optimization iteration, leading to unrealistic results.
{Since the rendered image is realistic at the last generation, a high level of noise will introduce large randomness.}
As shown in Fig.~\ref{fig-method-annealed-time-schedule}, the generated images become more different from the original one when a higher level of noise is added.

Based on this observation, we propose an annealed time schedule for optimization.
We linearly decrease $t_\text{max}$ as follows:
\begin{equation}
    t_\text{max} = t_1 -  (t_1 - t_0) \frac{n}{N},
\end{equation}
where $t_0$ and $t_1$ are the minimum and the maximum value of $t_\text{max}$, respectively. 
$n$ and $N$ are the current and the total steps of the optimization, respectively.

\noindent{\textbf{Reconstruction.}}
With images generated at specific poses and random poses, we optimize the NeRF using the reconstruction loss.
We compute the reconstruction loss for specific poses as follows:
\begin{equation}
    \mathcal{L}_{s}(\pmb{I}_s, \hat{\pmb{I}}_s) = \mathcal{L}_\text{LPIPS}(\pmb{I}_s, \hat{\pmb{I}}_s) + \mathcal{L}_\text{L1}(\pmb{I}_s, \hat{\pmb{I}}_s),
\end{equation}
where $\mathcal{L}_\text{LPIPS}(\cdot,\cdot)$ and $\mathcal{L}_\text{L1}(\cdot,\cdot)$ are the perceptual loss~\cite{zhang2018unreasonable} and the L1 loss, respectively.

We also regularize the NeRF at randomly sampled viewpoints. 
Similar to $\mathcal{L}_\text{s}$, the regularization loss $\mathcal{L}_\text{r}$ is defined as:
\begin{equation}
    \mathcal{L}_{r}(\pmb{I}_r, \hat{\pmb{I}}_r) = \mathcal{L}_\text{LPIPS}(\pmb{I}_r, \hat{\pmb{I}}_r) + \mathcal{L}_\text{L1}(\pmb{I}_r, \hat{\pmb{I}}_r).
\end{equation}

Totally, we optimize the NeRF with the following reconstruction loss:
\begin{equation}
    \mathcal{L}_\text{total} = \lambda_s \mathcal{L}_s + \lambda_r \mathcal{L}_r + \lambda_a \mathcal{L}_a,
\end{equation}
where $\lambda_s$, $\lambda_r$, and $\lambda_a$ are balancing weights for losses $\mathcal{L}_s$, $\mathcal{L}_r$, and $\mathcal{L}_a$. $\mathcal{L}_a$ is the additional geometry regularizaition of NeRF used in DreamFusion.

\subsection{Implementation}
We implement the proposed method based on the ThreeStudio\footnote{https://github.com/threestudio-project/threestudio} framework.
We use the improved Mip-NeRF 360~\cite{barron2022mip} with neural hashing encoding~\cite{muller2022instant} as the 3D representation.
We use the Stable Diffuson v1.5~\cite{metzer2022latent} along with ControlNet v1.1~\cite{zhang2023adding} as our sketch-conditioned guidance.
We set $t_\text{min}$, $t_0$, and $t_1$ to 0.02, 0.5, and 0.98, respectively. 
We optimize the proposed method for 25000 iterations for each object (i.e., sketch images and a prompt).
We use the Adam optimizer with an initial learning rate of 0.01.
We set $\lambda_s$ and $\lambda_r$ to 10.0 and 1.0, respectively.
We set $\lambda_a$ to 1.0 after 20000 iterations, otherwise 0.
We found that the strategy of $\lambda_a$ is critical to generate high-quality geometry.
Our method takes around 2 hours to generate a 3D object on a single NVIDIA RTX 3090 GPU.
\section{Experiments}
We first introduce the datasets (Sec.~\ref{Datasets}) and metrics (Sec.~\ref{Evaluation Metrics}) for performance evaluation. 
We then provide comparisons with several baseline methods (Sec.~\ref{Baselines} and ~\ref{Results}). 
Finally, we show the ablative results (Sec.~\ref{Ablation Study}).

\subsection{Datasets}
\label{Datasets}
Since the multi-view sketch dataset does not exist, we collect two kinds of multi-view sketch datasets (i.e., the OmniObject3D-Sketch dataset and the THuman-Sketch dataset) to evaluate the proposed method.

\noindent\textbf{OmniObject3D-Sketch.}
We collect 20 categories of objects from the OmniObject3D dataset~\cite{wu2023omniobject3d}.
In particular, we use 24 rendered images of each object provided by~\cite{wu2023omniobject3d}.
The sketch images are extracted by performing the HED boundary detector~\cite{xie2015holistically} on these rendered images, while the prompt is obtained by performing the image caption model (i.e., BLIP-2~\cite{li2023blip}) on a rendered image for each object.
The OmniObject3D-Sketch dataset covers a wide range of daily 3D objects from the real world.

\noindent\textbf{THuman-Sketch.}
We collect 20 different human meshes from the THuman 3.0 dataset~\cite{su2022deepcloth}.
After rendering 16 images with different poses located at the upper hemisphere of each mesh, we employ the HED boundary detector to obtain the sketch of each rendered image.
The prompts of all humans are fixed to “\textit{a DSLR photo of a human}”.
The Thuman-Sketch dataset contains several human postures which are hard to describe with texts but easy to depict with sketches.


\begin{figure*}[hbpt]
    \centering
    \begin{subfigure}{0.16\linewidth}
        \begin{minipage}[t]{\linewidth}
            \includegraphics[width=\linewidth]{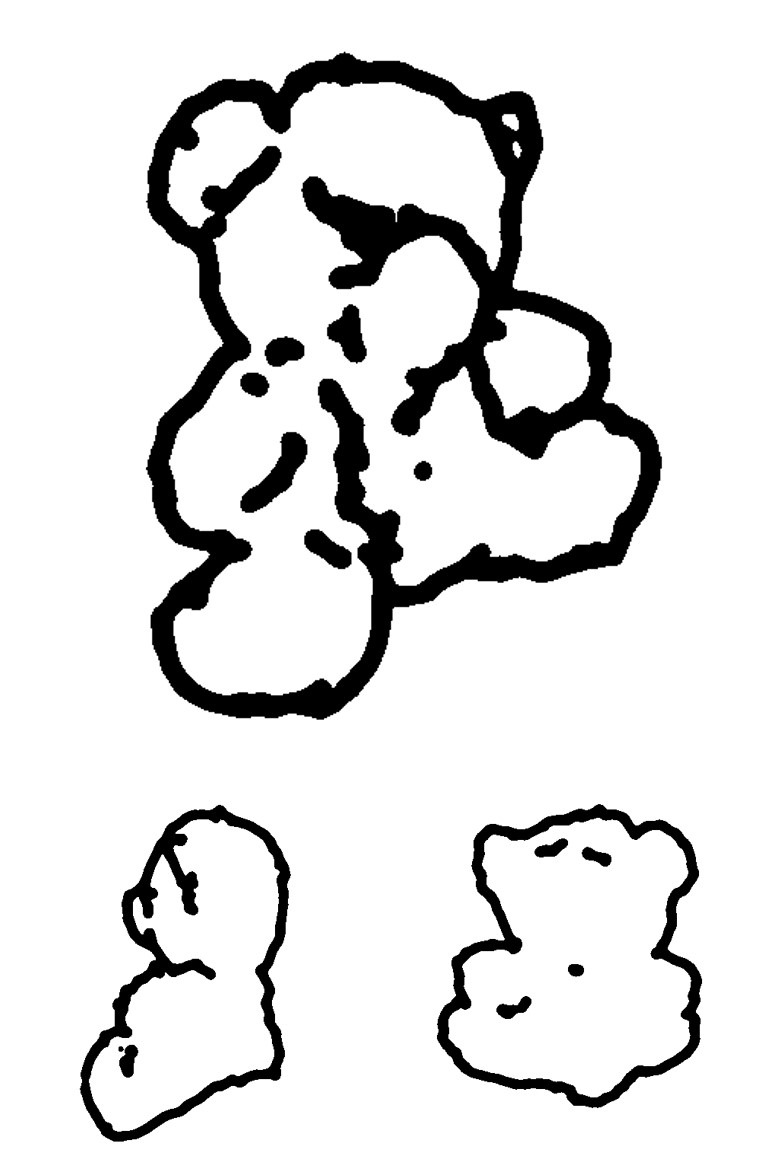}
            \includegraphics[width=\linewidth]{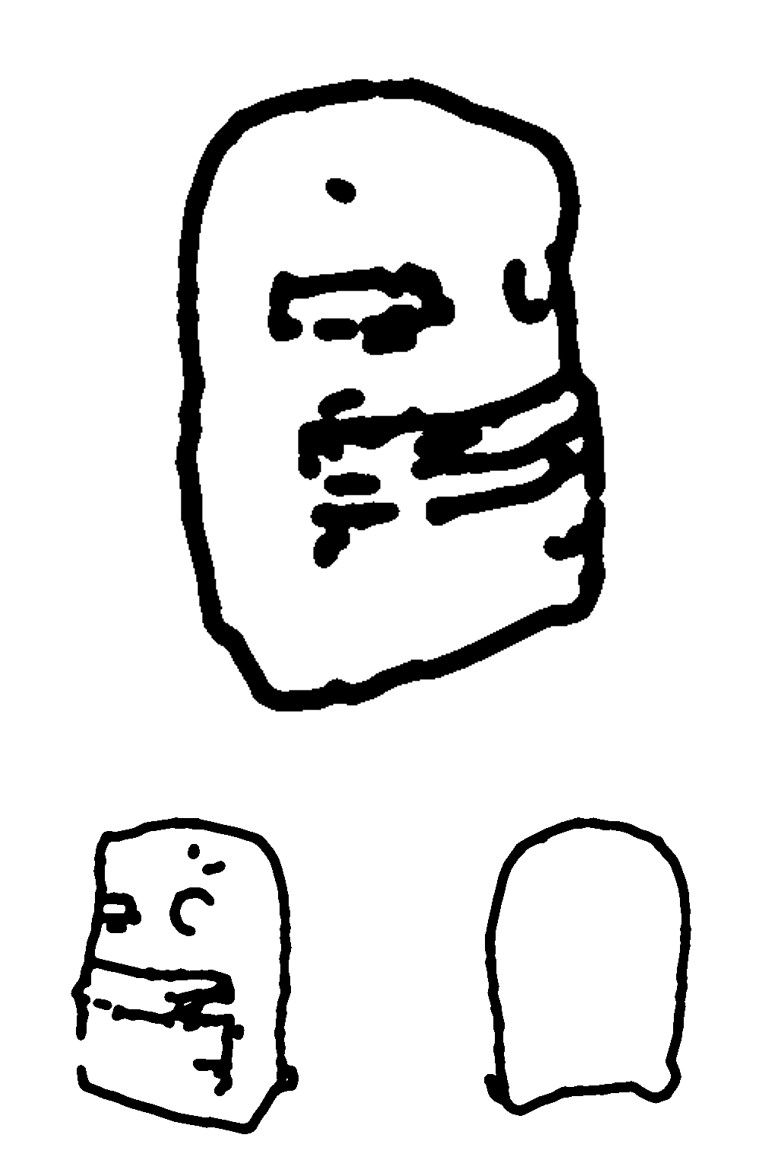}
            \includegraphics[width=\linewidth]{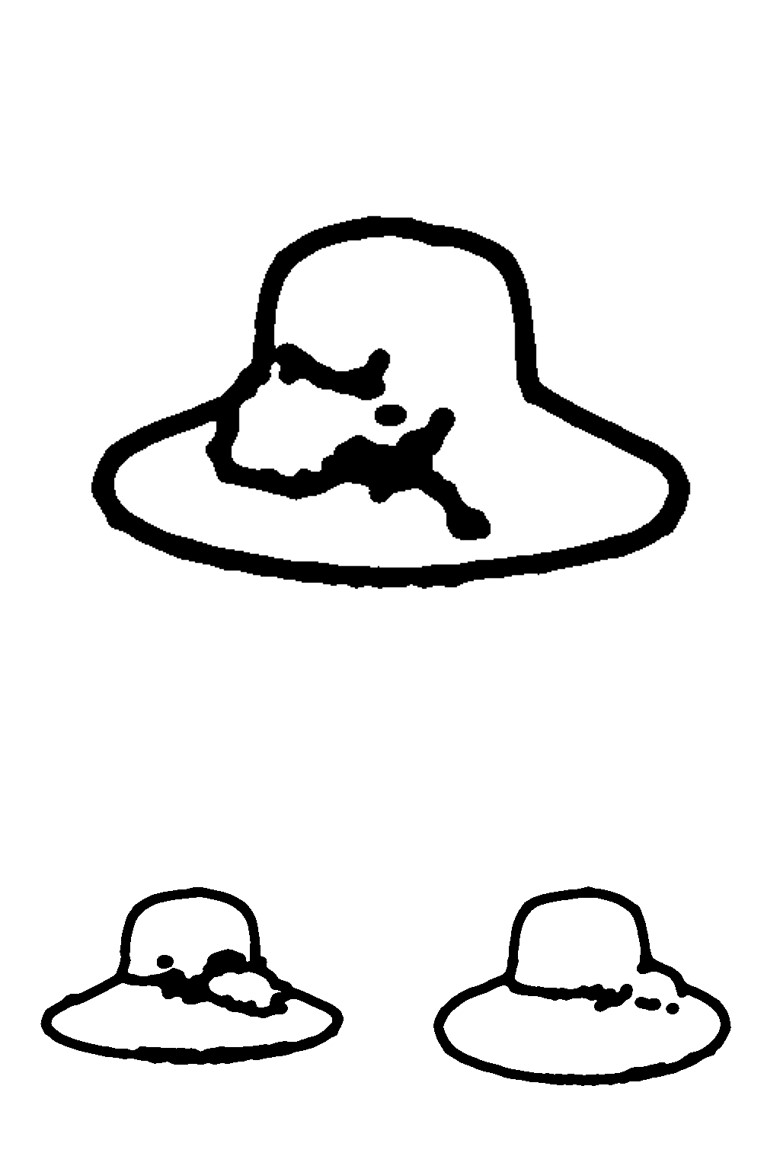}
        \end{minipage}
        \caption{Input Sketches}
    \end{subfigure}
    \begin{subfigure}{0.16\linewidth}
        \begin{minipage}[t]{\linewidth}
            \includegraphics[width=\linewidth]{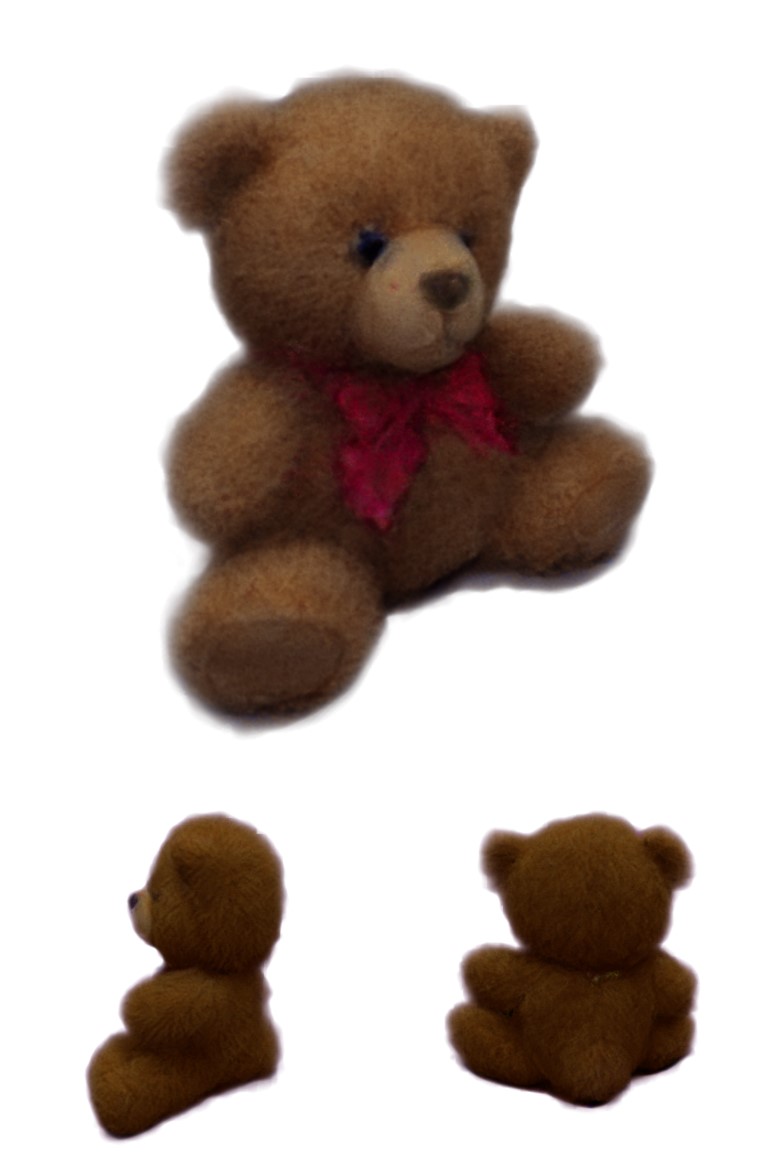}
            \includegraphics[width=\linewidth]{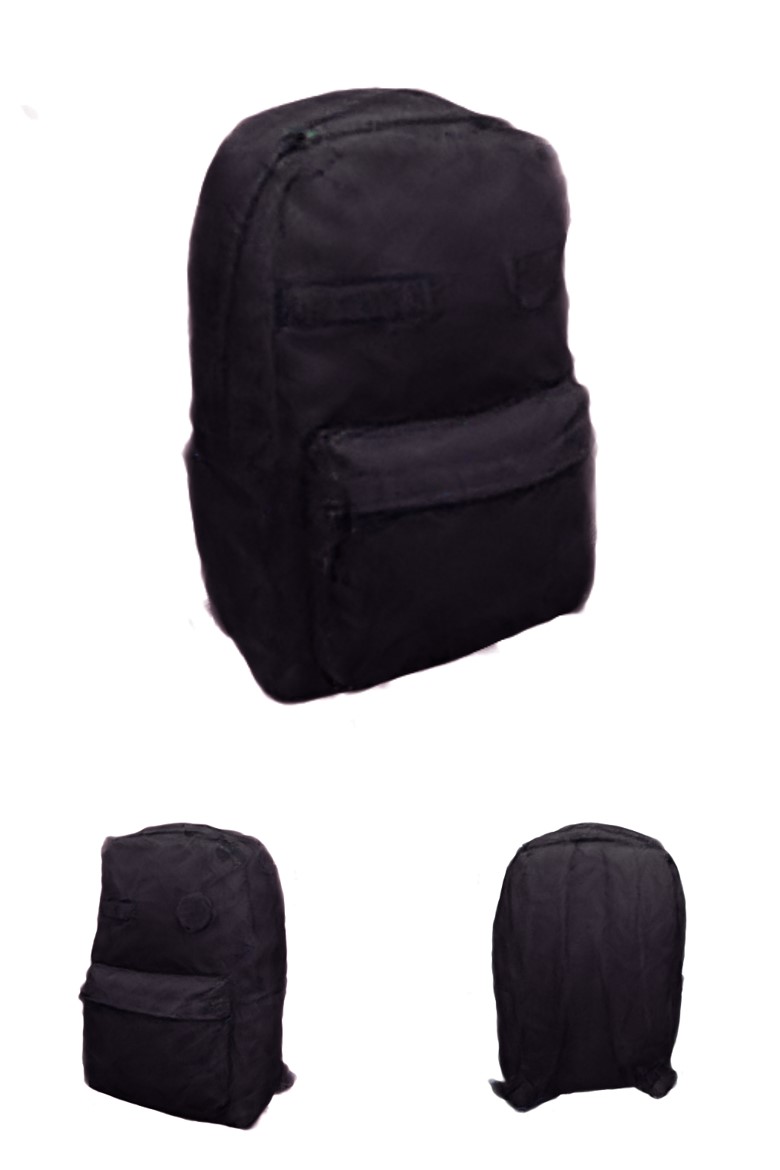}
            \includegraphics[width=\linewidth]{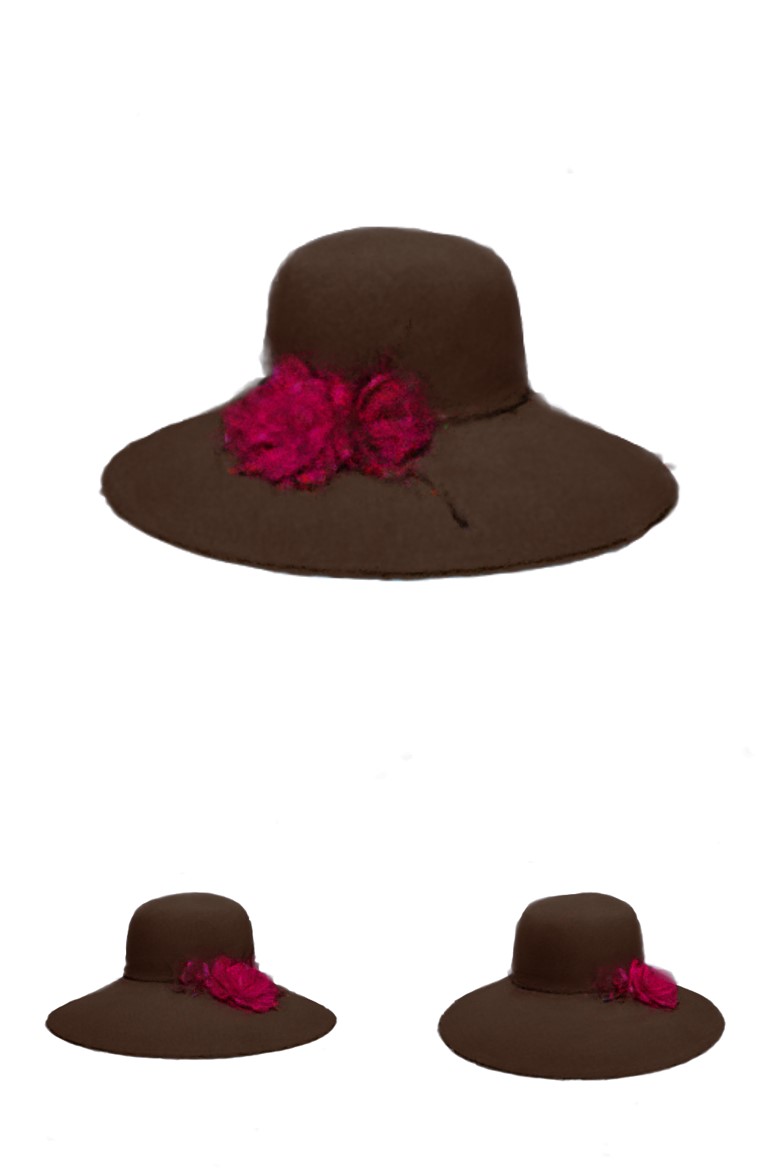}
        \end{minipage}
        \caption{Ours}
    \end{subfigure}
    \begin{subfigure}{0.16\linewidth}
        \begin{minipage}[t]{\linewidth}
            \includegraphics[width=\linewidth]{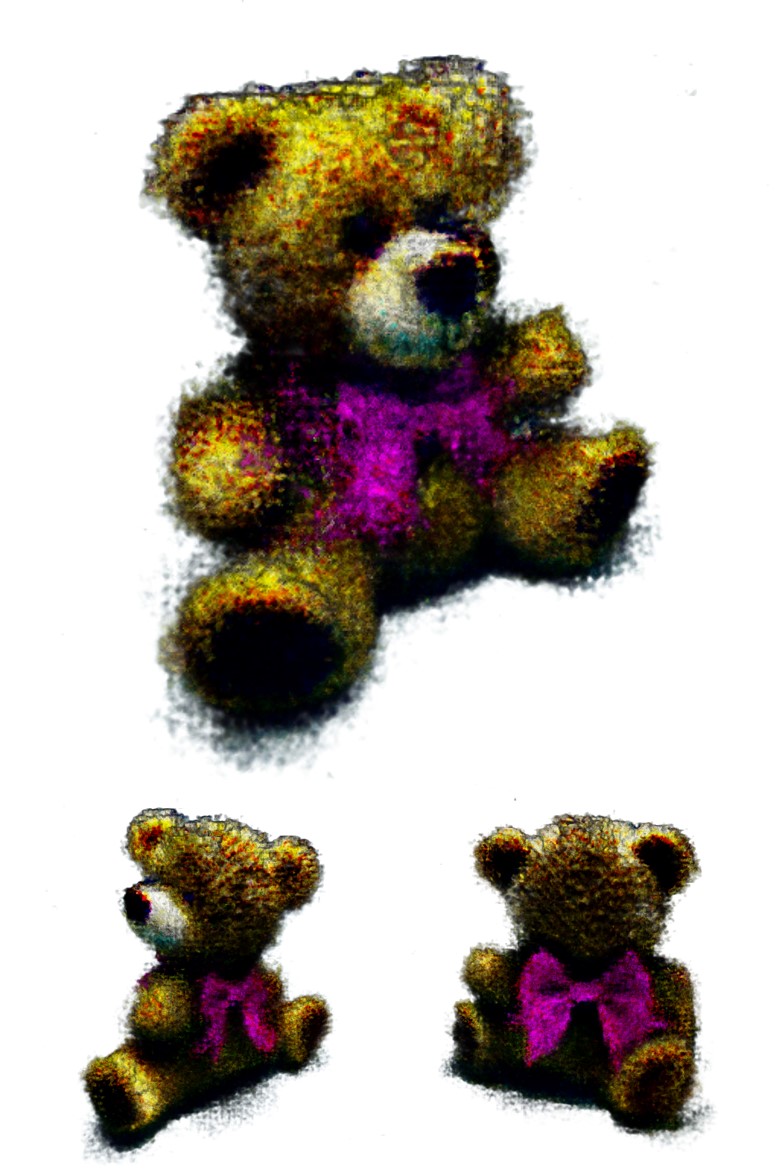}
            \includegraphics[width=\linewidth]{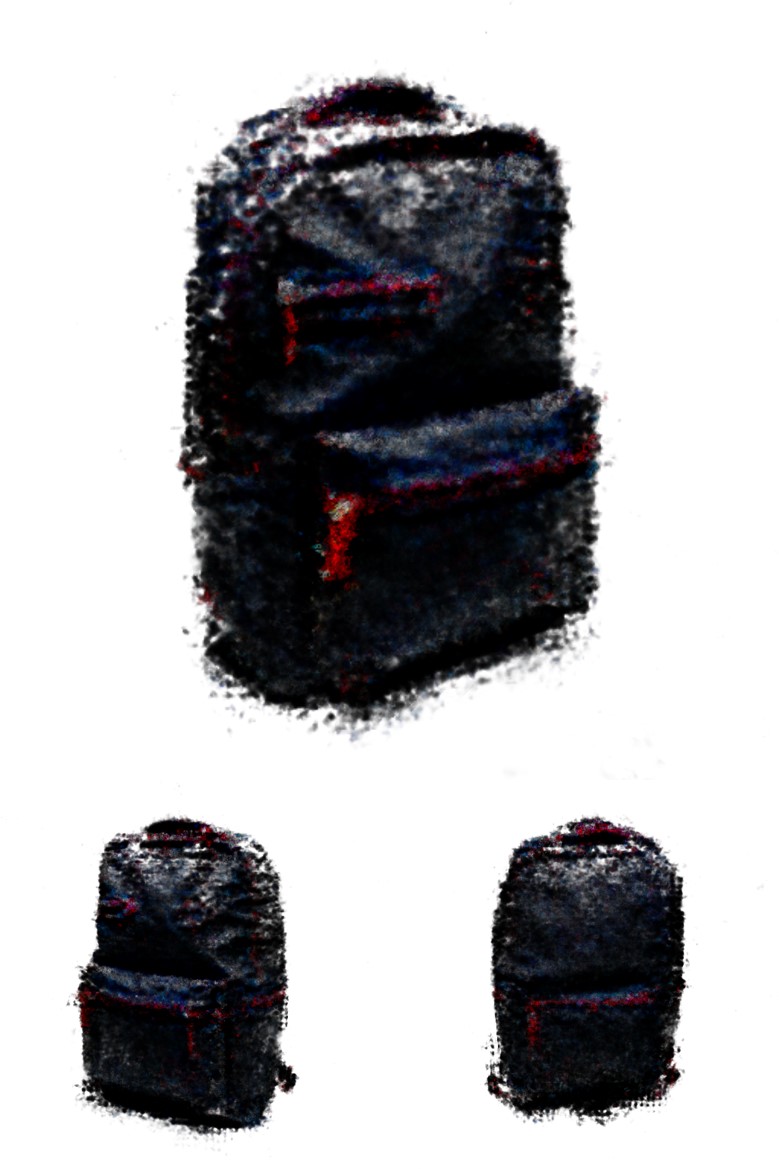}
            \includegraphics[width=\linewidth]{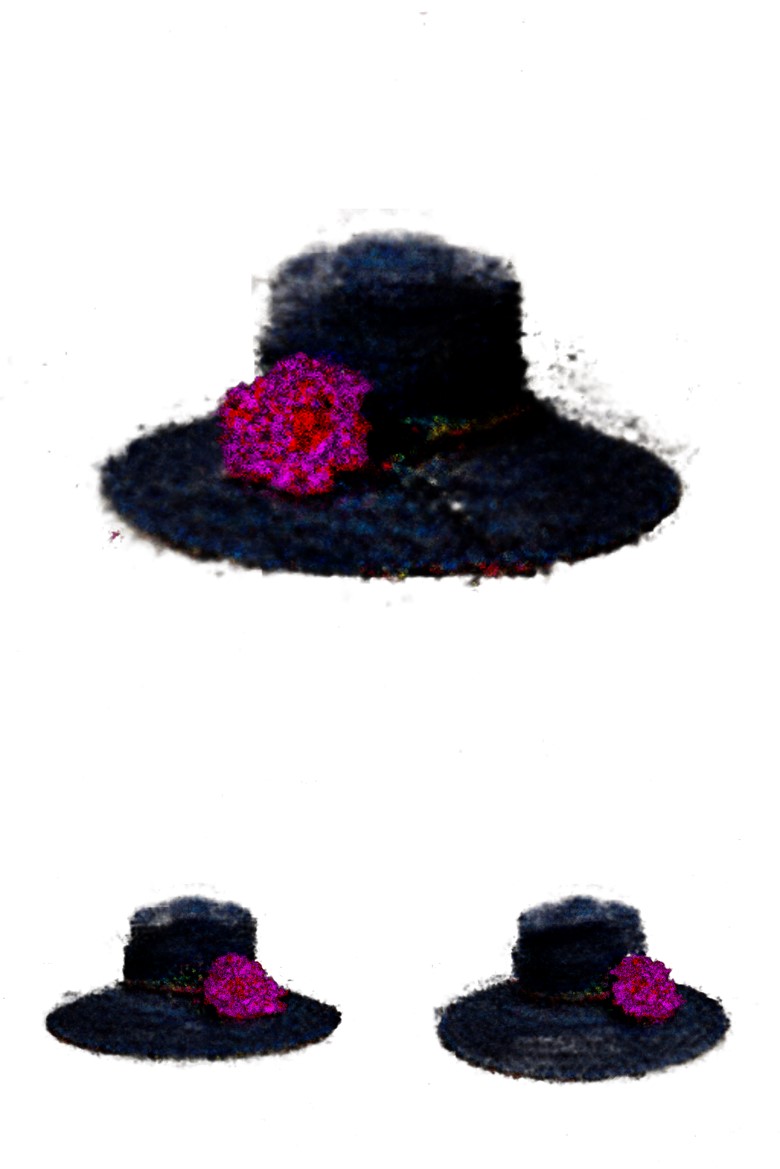}
        \end{minipage}
        \caption{C-DreamFusion}
    \end{subfigure}
    \begin{subfigure}{0.16\linewidth}
        \begin{minipage}[t]{\linewidth}
            \includegraphics[width=\linewidth]{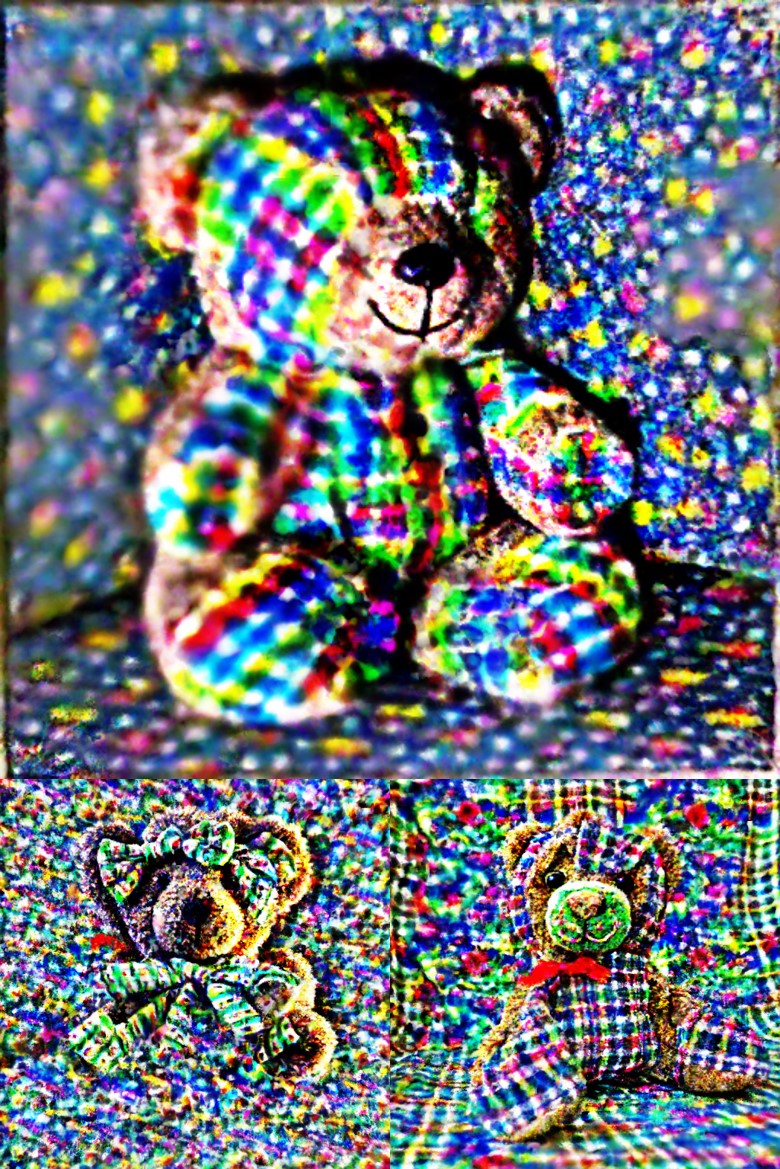}
            \includegraphics[width=\linewidth]{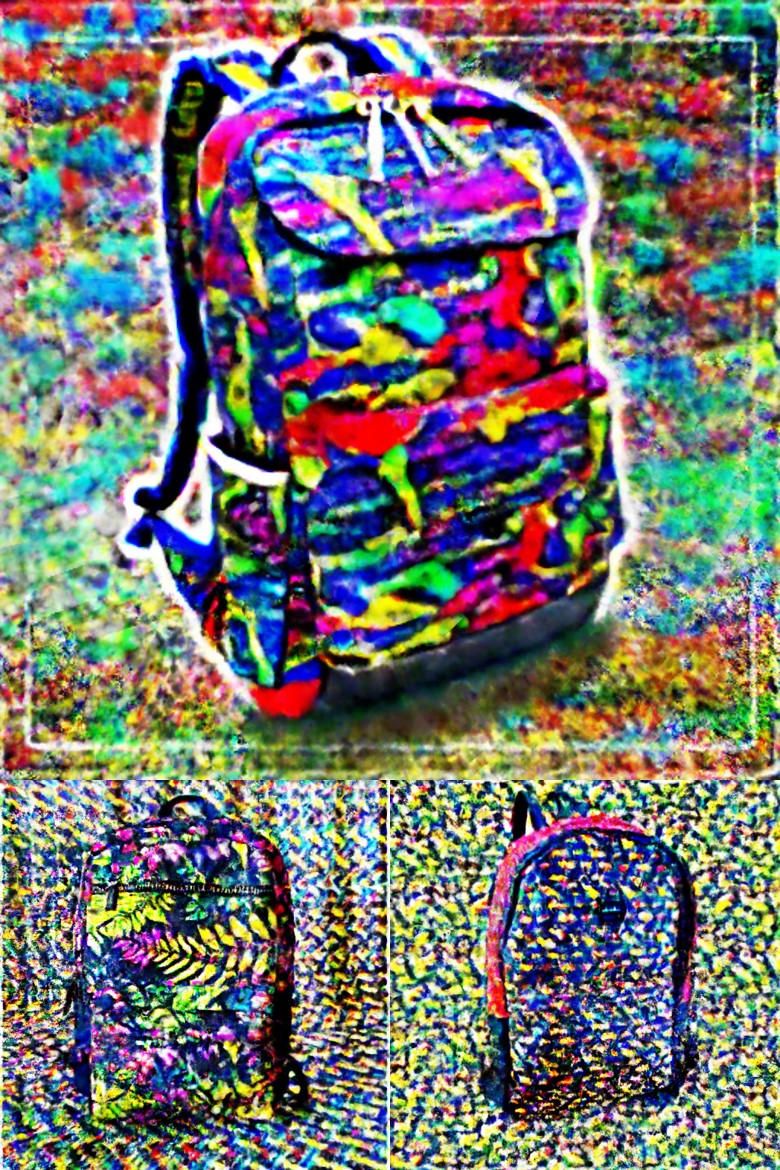}
            \includegraphics[width=\linewidth]{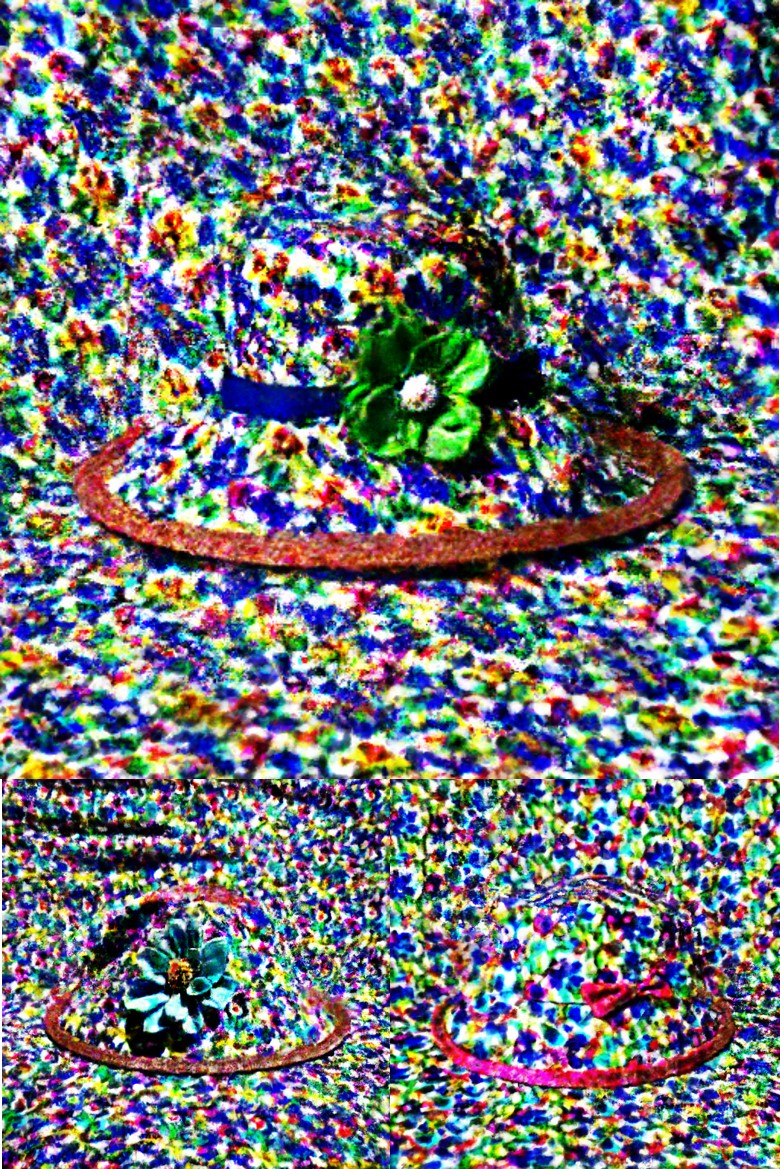}
        \end{minipage}
        \caption{C-ProlificDreamer}
    \end{subfigure}
    \begin{subfigure}{0.16\linewidth}
        \begin{minipage}[t]{\linewidth}
            \includegraphics[width=\linewidth]{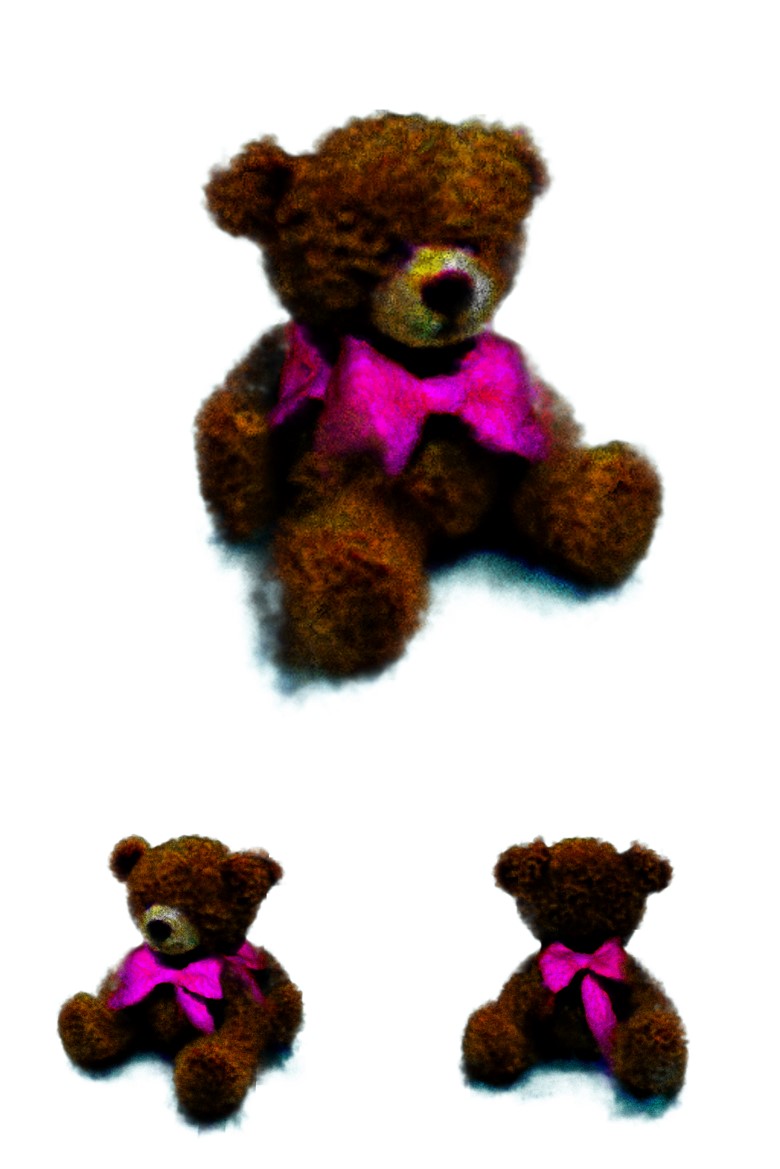}
            \includegraphics[width=\linewidth]{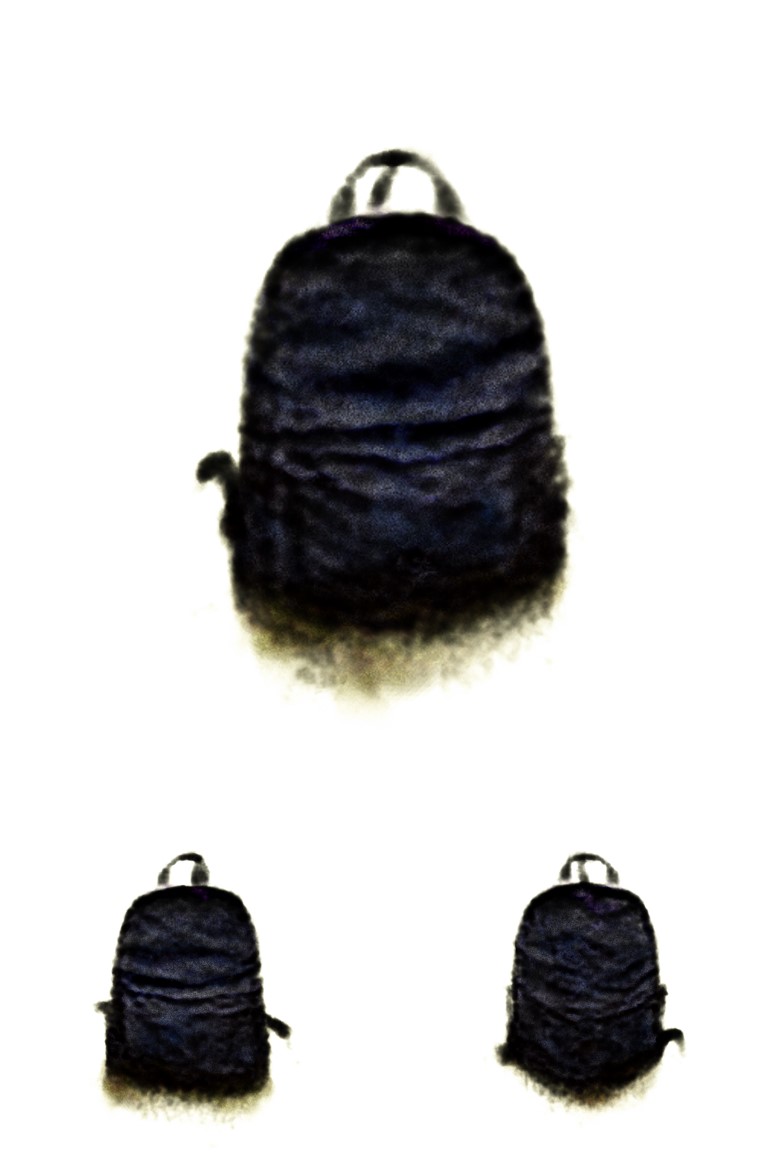}
            \includegraphics[width=\linewidth]{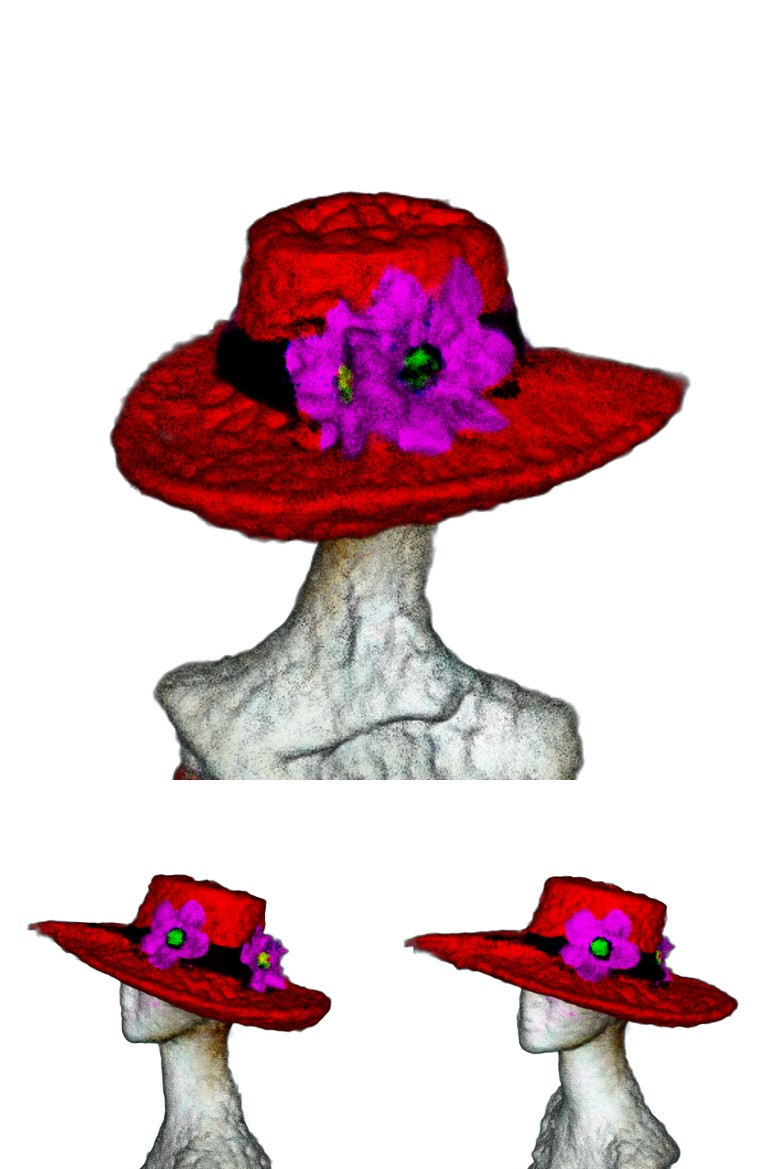}
        \end{minipage}
        \caption{DreamFusion~\cite{poole2022dreamfusion}}
    \end{subfigure}
    \begin{subfigure}{0.16\linewidth}
        \begin{minipage}[t]{\linewidth}
            \includegraphics[width=\linewidth]{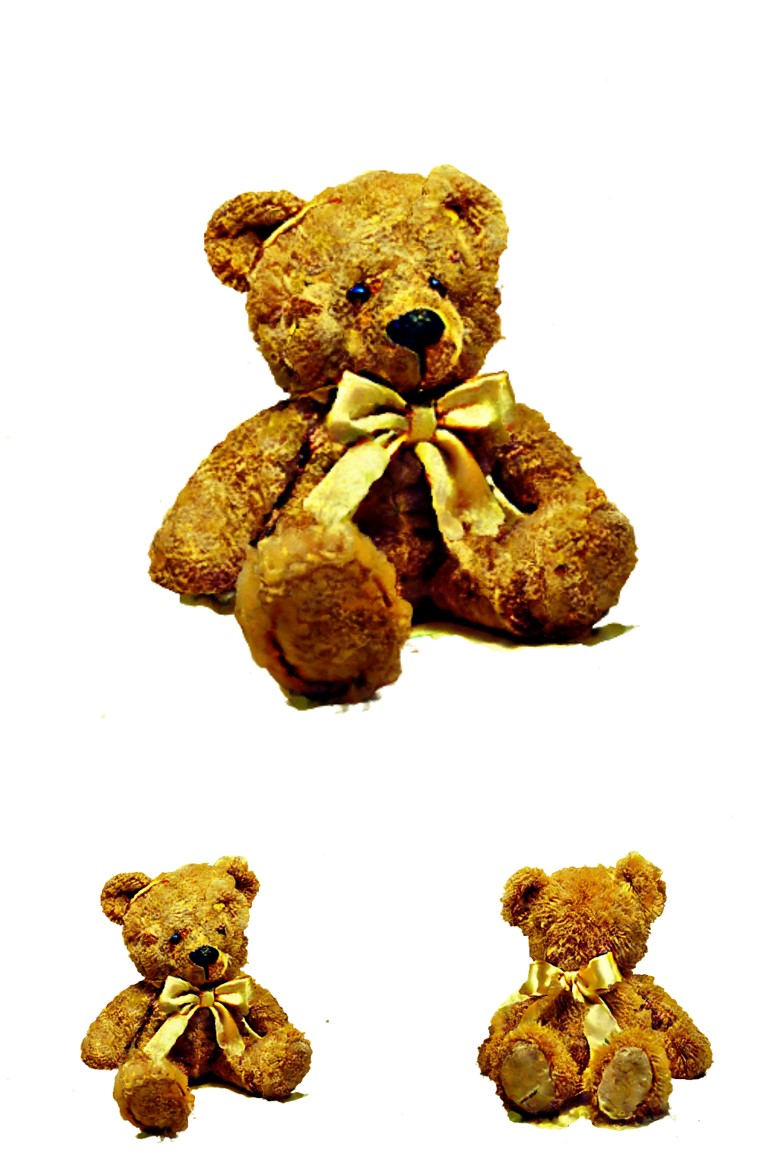}
            \includegraphics[width=\linewidth]{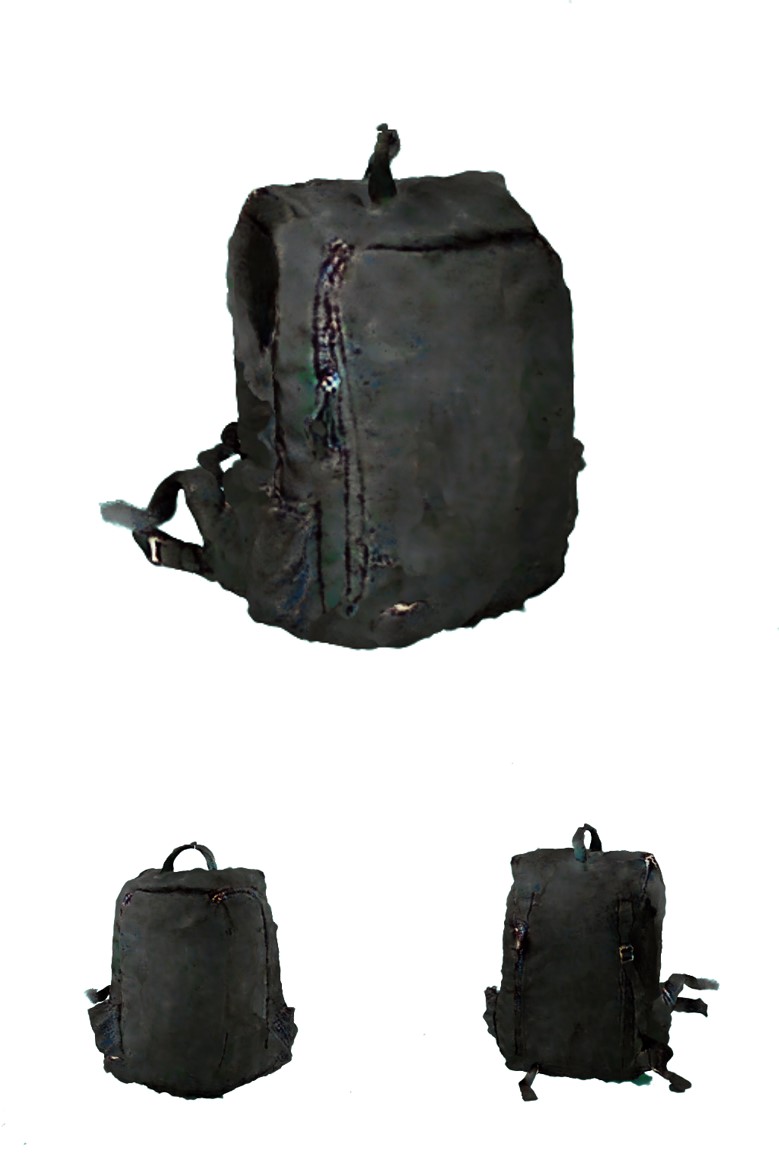}
            \includegraphics[width=\linewidth]{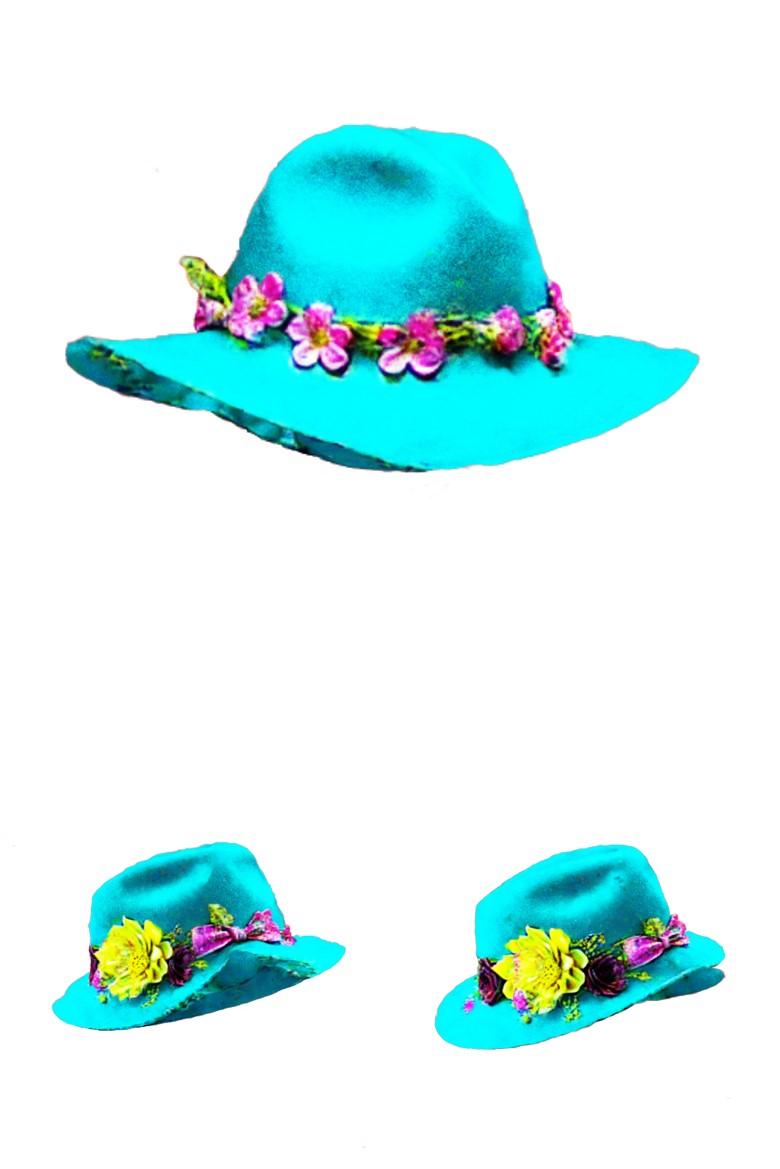}
        \end{minipage}
        \caption{ProlificDreamer~\cite{wang2023prolificdreamer}}
    \end{subfigure}
    \caption{
        Qualitative comparisons on 3 different objects with four baseline methods.
        Results clearly indicate that our method produces better consistent and high-fidelity 3D objects with multi-view sketch control.
    }
    \label{fig-results-qualitative-comparisons}
    \vspace{-0.4cm}
\end{figure*}

\begin{figure*}[hbpt]
    \centering
    \begin{subfigure}{0.24\linewidth}
        \begin{minipage}[c]{\linewidth}
            \includegraphics[width=1.00\linewidth]{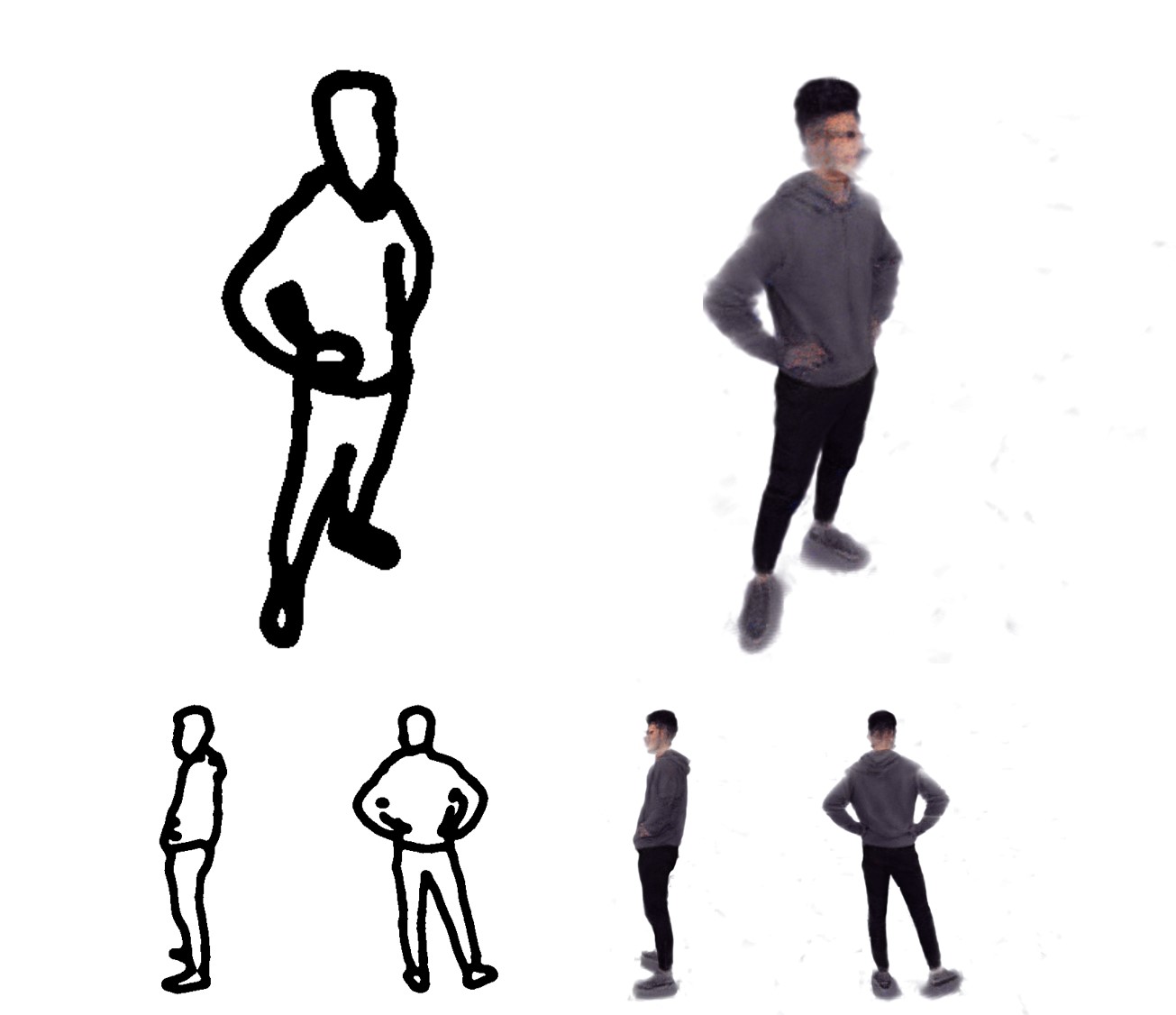}
        \end{minipage}
    \end{subfigure}
    \begin{subfigure}{0.24\linewidth}
        \begin{minipage}[c]{\linewidth}
            \includegraphics[width=1.00\linewidth]{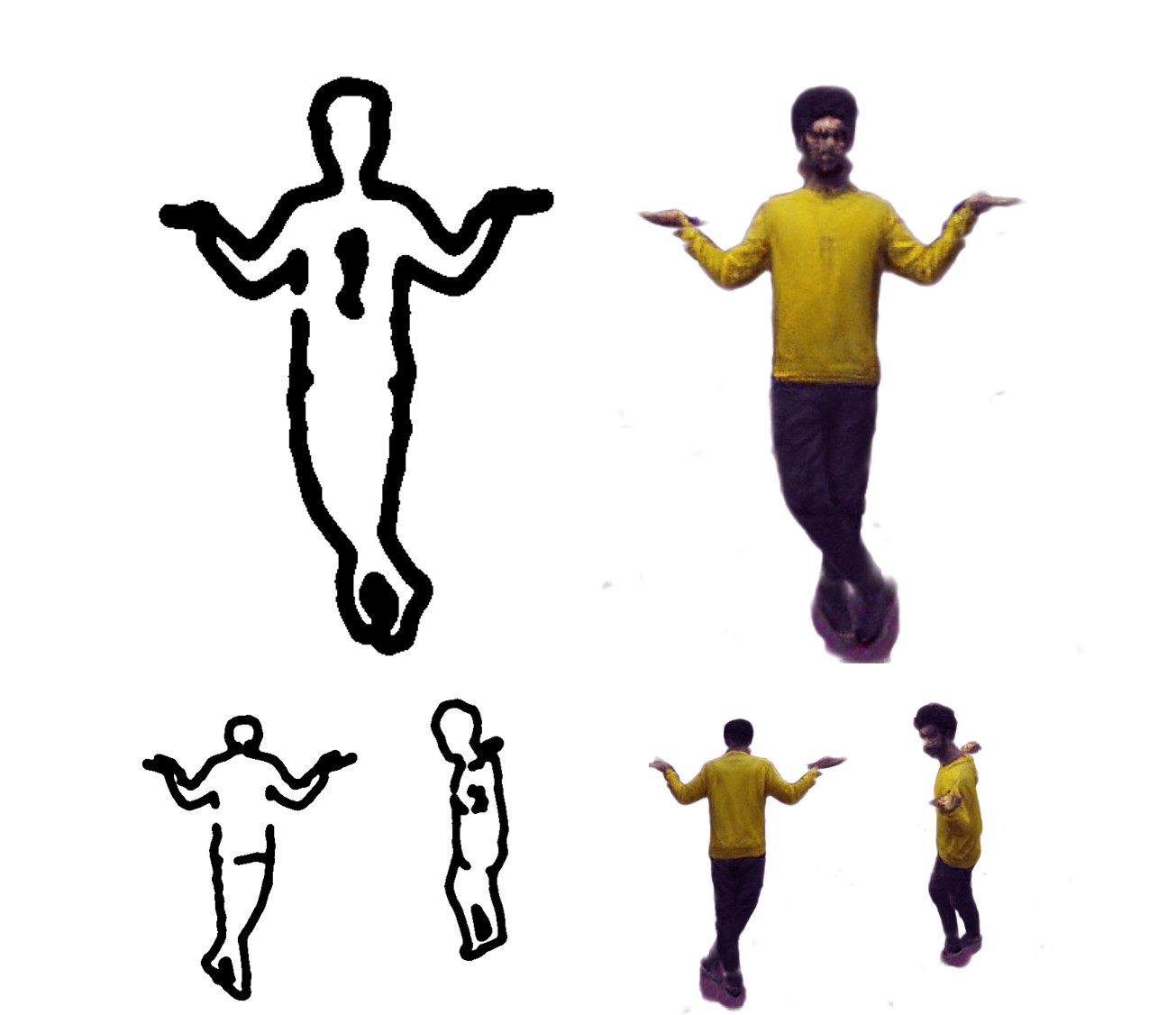}
        \end{minipage}
    \end{subfigure}
    \begin{subfigure}{0.24\linewidth}
        \begin{minipage}[c]{\linewidth}
            \includegraphics[width=1.00\linewidth]{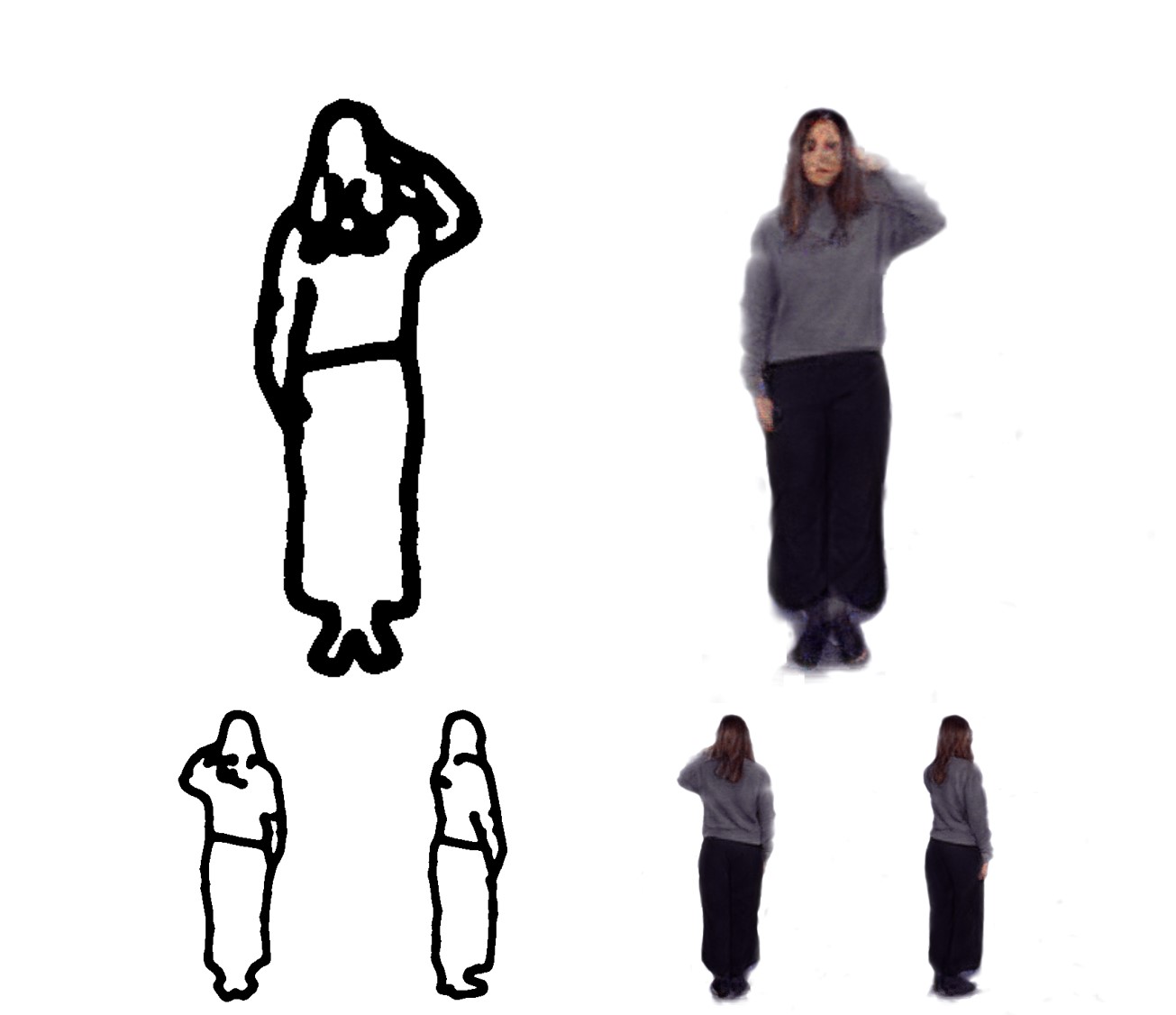}
        \end{minipage}
    \end{subfigure}
    \begin{subfigure}{0.24\linewidth}
        \begin{minipage}[c]{\linewidth}
            \includegraphics[width=1.00\linewidth]{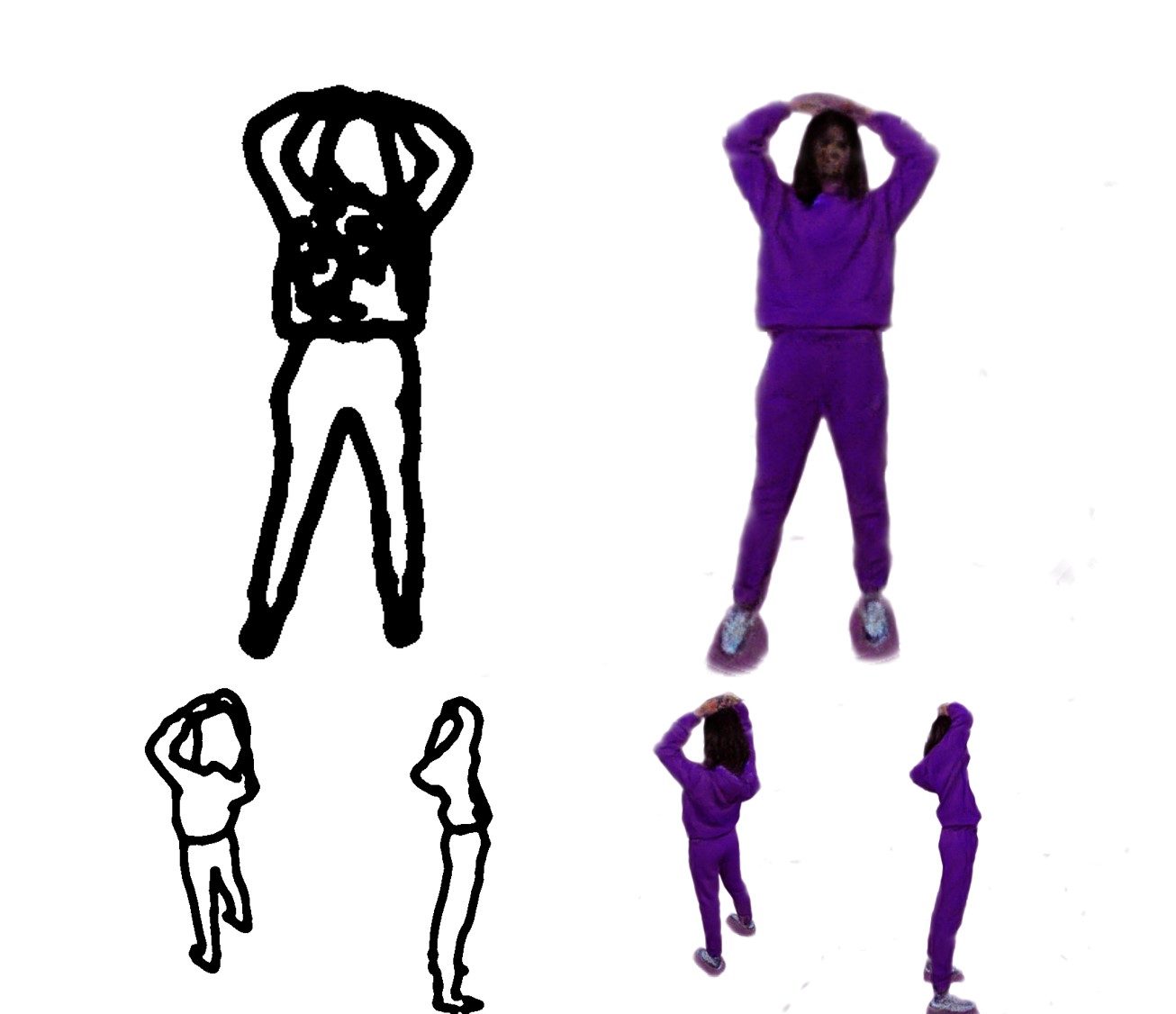}
        \end{minipage}
    \end{subfigure}
    \caption{
        Visual results of human generation by our method on the THuman-Sketch dataset. We show four generated 3D humans. For each 3D human, we show the input mult-view sketches (\textit{left}), and the generated 3D human (\textit{right}).
    }
    \label{fig-results-human-generation}
    \vspace{-0.2cm}
\end{figure*}

\begin{table*}[htbp]
  \centering
  \caption{Quantitative comparisons on the OmniObject3D-Sketch dataset. (Note that RP denotes R-Precision)}
    \begin{tabular}{l|c|cc|ccc}
    \toprule
          & \multirow{2}[2]{*}{Sketch Input} & \multicolumn{2}{c|}{Sketch Similarity} & \multicolumn{3}{c}{Text Alignment} \\
          &       & CD $\downarrow$   & HD $\downarrow$    & RP CLIP B/32 $\uparrow$ & RP CLIP B/16 $\uparrow$ & RP CLIP L/14 $\uparrow$ \\
    \midrule
    DreamFusion~\cite{poole2022dreamfusion}         & ×                 & 0.0411            & 0.1876            & 0.7511            & 0.7753            & 0.7972 \\
    ProlificDreamer~\cite{wang2023prolificdreamer}  & ×                 & 0.0438            & 0.1784            & 0.7687            & 0.7907            & 0.8457 \\
    C-DreamFusion                                   & \checkmark        & 0.0148            & 0.1478            & 0.7039            & 0.7171            & 0.7848 \\
    C-ProlificDreamer                               & \checkmark        & 0.0458            & 0.3273            & 0.0675            & 0.0531            & 0.0491 \\
    Sketch2NeRF (Ours)                              & \checkmark        & \textbf{0.0091}   & \textbf{0.1233}   & \textbf{0.7821}   & \textbf{0.8110} & \textbf{0.8597} \\
    \bottomrule
    \end{tabular}%
    \vspace{-0.4cm}
  \label{tab-quantitative-comparison-omniobject3d-sketch}%
\end{table*}%

\subsection{Evaluation Metrics}
\label{Evaluation Metrics}

\noindent\textbf{Sketch Similarity}
We evaluate the similarity between the input sketches and the generated object sketches.
Since sketches can be considered as image edges, we introduce two edge-based metrics (i.e., \textit{chamfer distance} (CD)~\cite{yuan2018pcn}, and \textit{hausdorff distance} (HD)~\cite{huttenlocher1993comparing}) to evaluate the sketch similarity.

\noindent\textbf{Text Alignment}
We evaluate the alignment of generated 3D objects with their prompt texts using the CLIP R-Precision metric~\cite{jain2022zero,poole2022dreamfusion}.
Specifically, we render images of size $512 \times 512$ for each generated object, where its pose is randomly sampled on the upper hemisphere.
We use all prompts from the OmniOBject3D-Sketch dataset.
The CLIP R-Precision measures the retrieval accuracy of the rendered image and the prompts.
We provided the CLIP R-Precision metrics using three different CLIP models~\cite{radford2021learning} (i.e., ViT-B/32, ViT-B/16, and ViT-L-14).

\subsection{Baselines}
\label{Baselines}

We include two types of baseline methods for comparison.
We first compare our method to DreamFusion~\cite{poole2022dreamfusion} and ProlificDreamer~\cite{wang2023prolificdreamer}, which are proposed for text-conditioned 3D object generation.
DreamFusion is the first to distill knowledge from 2D diffusion model to 3D objects using the SDS technique, while ProlificDreamer achieves state-of-the-art performance in the text-conditioned 3D generation.
To achieve multi-view sketch control, we replace the Stable Diffusion used in DreamFusion and ProlificDreamer with ControlNet. We refer to the modified versions of DreamFusion and ProlificDreamer as C-DreameFusion and C-ProlificDreamer, respectively.
Note that we use the implementation of DreamFusion in the ThreeStudio framework, since the official codebase is not released.

\subsection{Results}
\label{Results}

\noindent\textbf{Qualitative Comparisons.}
Fig.~\ref{fig-results-qualitative-comparisons} shows the visual results of our method and other competitive baseline methods.
The results demonstrate that our method can generate a wide range of plausible 3D objects.
In addition, the generation of our method can be controlled by the input sketches.
The generated 3D objects resemble the specific sketches in detail, e.g., the zipper of the backpack, and the flower position on the hat (as shown in Fig.~\ref{fig-results-qualitative-comparisons}).
In contrast, the generated 3D objects of DreamFusion and ProlificDream are stochastic and cannot be constrained by multi-view sketches.
Although C-DreamFusion and C-ProlificDream can be guided by multi-view sketches, C-DreamFusion produces more noise on the edge of 3D objects, while C-ProlificDreamer produces an unpleasant background.
This demonstrates the fine-grained controllability of our method in 3D generation.

Fig.~\ref{fig-results-human-generation} shows the human generation results produced by our method on the THuman-Sketch dataset. 
Human generation with only text conditions is highly challenging due to the large variation of posture space in a generation. 
In practice, we found that DreamFusion and ProlificDreamer failed to synthesize humans, resulting in empty or low-quality 3D shapes.
However, our method can synthesize high-quality humans with various postures and a fixed simple prompt text (Fig.~\ref{fig-results-human-generation}).
The generated humans not only resemble the given sketches but also have high-fidelity clothes.
This demonstrates the promising performance of our method on human generation without explicit geometry constraints.

\begin{figure}[t]
    \centering
    \begin{subfigure}{0.48\linewidth}
        \begin{minipage}[t]{\linewidth}
            \includegraphics[width=0.49\linewidth]{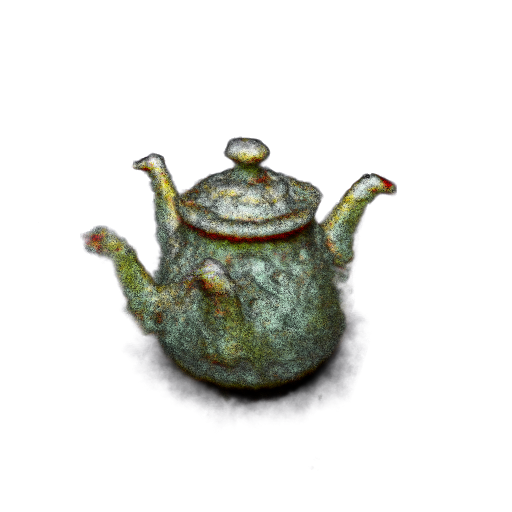}
            \includegraphics[width=0.49\linewidth]{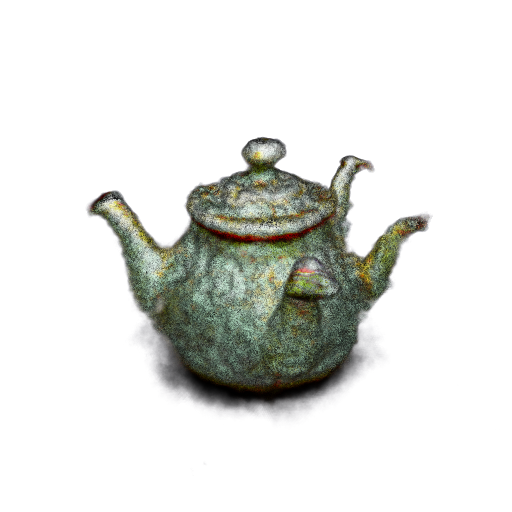}\\
            \includegraphics[width=0.49\linewidth]{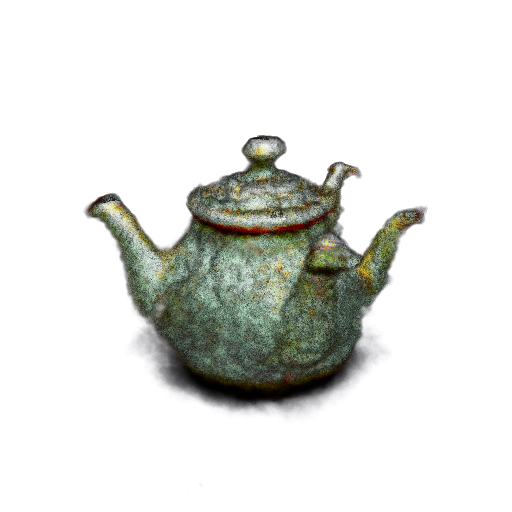}
            \includegraphics[width=0.49\linewidth]{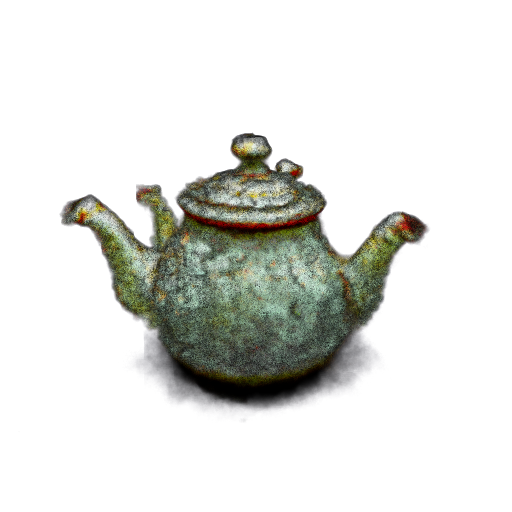}\\
            \includegraphics[width=0.49\linewidth]{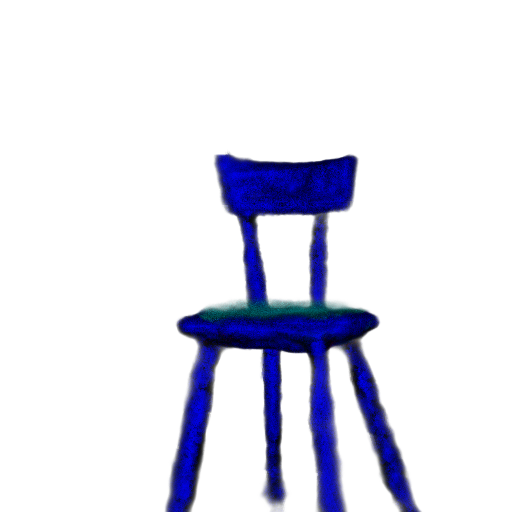}
            \includegraphics[width=0.49\linewidth]{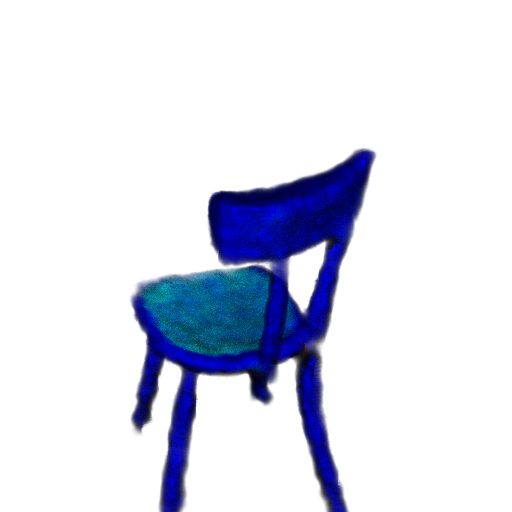}\\
            \includegraphics[width=0.49\linewidth]{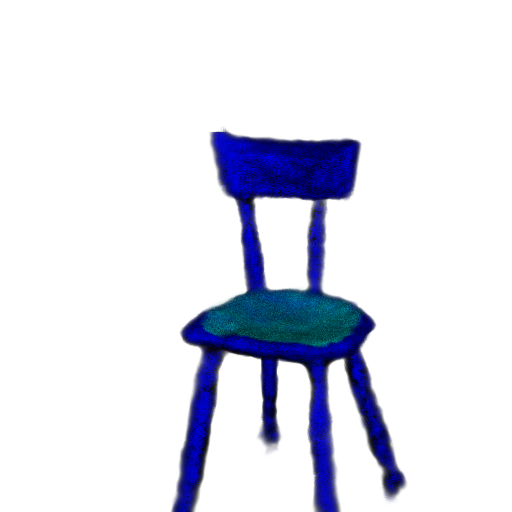}
            \includegraphics[width=0.49\linewidth]{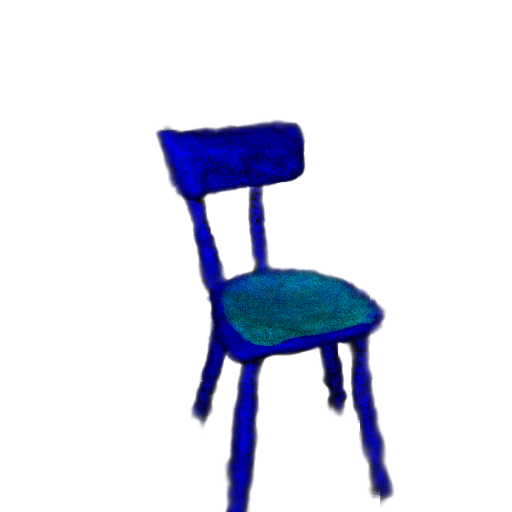}
        \end{minipage}
        \caption{DreamFusion~\cite{poole2022dreamfusion}}
    \end{subfigure}
    \begin{subfigure}{0.48\linewidth}
        \begin{minipage}[t]{\linewidth}
            \includegraphics[width=0.49\linewidth]{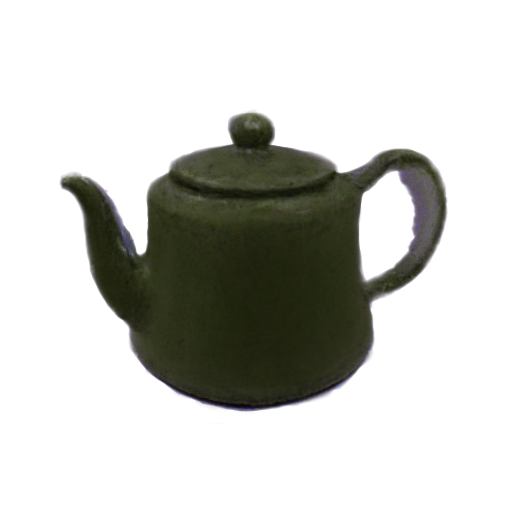}
            \includegraphics[width=0.49\linewidth]{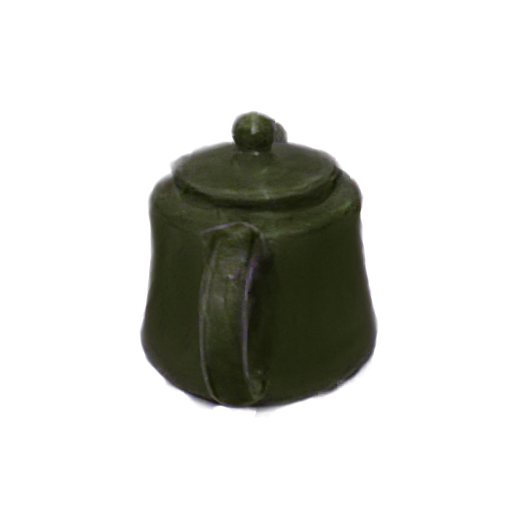}\\
            \includegraphics[width=0.49\linewidth]{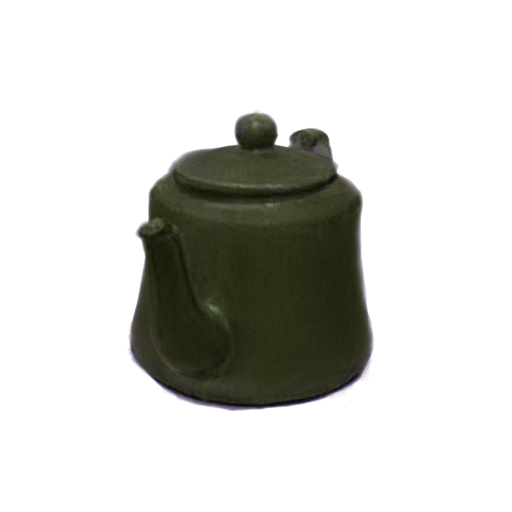}
            \includegraphics[width=0.49\linewidth]{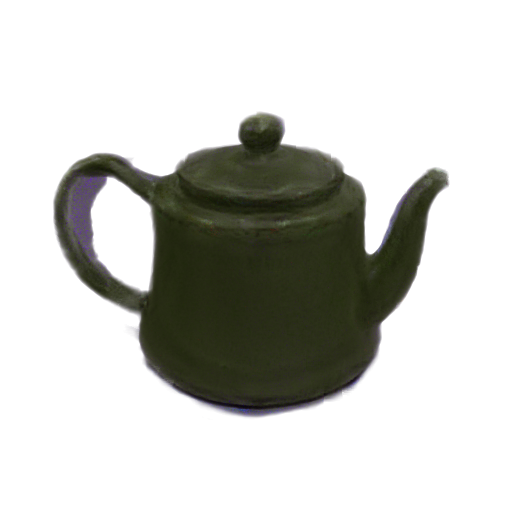}\\
            \includegraphics[width=0.49\linewidth]{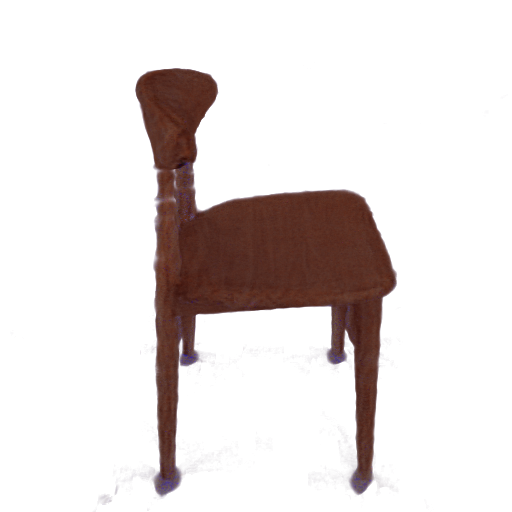}
            \includegraphics[width=0.49\linewidth]{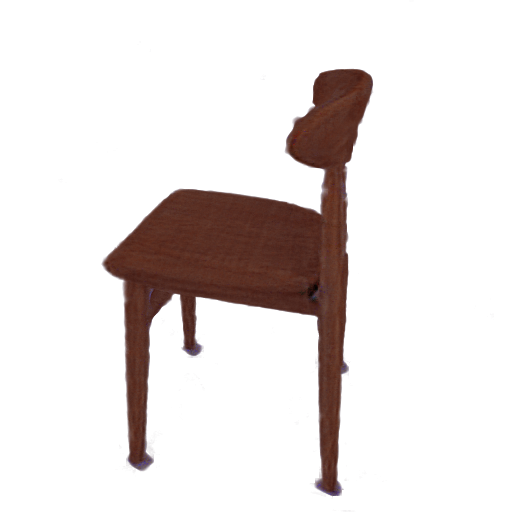}\\
            \includegraphics[width=0.49\linewidth]{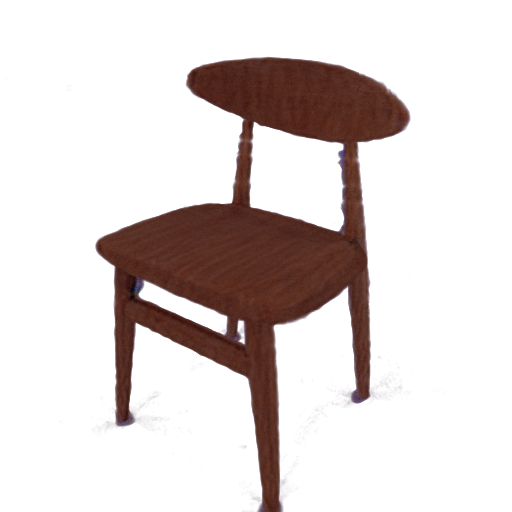}
            \includegraphics[width=0.49\linewidth]{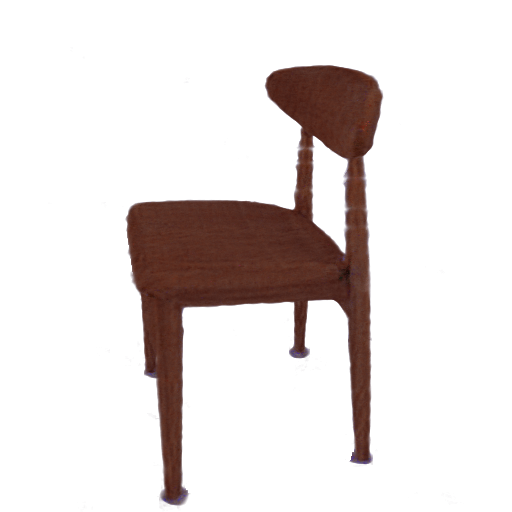}
        \end{minipage}
        \caption{Ours}
    \end{subfigure}
    \caption{
        Comparisons on hard cases with DreamFusion.
    }
    \label{fig-results-3d-consistency}
    \vspace{-0.4cm}
\end{figure}

Fig.~\ref{fig-results-3d-consistency} shows that our method can generate consistent 3D objects.
DreamFusion faces the \textit{Janus} problem, that is, the generated 3D object has multiple faces. For example, the teapot generated by DreamFusion has multiple spouts.
This is because the 2D generative diffusion model used by DreamFusion is unaware of 3D constraints.
With the multi-view sketch constraint, our method can alleviate the Janus problem.
In addition, we find that our method produces human-made regular 3D objects of higher quality than DreamFusion.
For example, DreamFusion can generate a chair with four legs but struggles to correctly arrange them at four corners.
With the sketch constraint, our method produces a regular chair that matches the real world.
This demonstrates that our method can generate objects with 3D consistency.

\noindent\textbf{Quantitative Comparisons.}
Tab.~\ref{tab-quantitative-comparison-omniobject3d-sketch} shows the metrics of sketch similarity and text alignment achieved by various methods.
Compared to text-to-3D approaches (i.e., DreamFusion and ProlificDreamer), our method achieves significantly better sketch similarity in terms of CD and HD.
Our method achieves a CD of 0.0091 which outperforms DreamFusion and ProlificDreamer by 0.032 and 0.034, respectively.
This demonstrates that our method can be controlled by given sketches.
In addition, our method achieves better text alignment than ProlificDreamer in terms of R-Precision CLIP with B/32, B/16, and L/14 ViT models.
Our method improves ProlificDream by 1.7\% in terms of R-Precision CLIP B/32.

Compared to the modified version of DreamFusion and ProlificDreamer, our method outperforms C-DreamFusion and C-ProlificDreamer by 0.0057 and 0.0367 in terms of CD, respectively.
Besides, our method outperforms C-DreamFusion by 0.0782 in terms of R-Precision CLIP B/32.
This is because the proposed synchronized generation and reconstruction method can effectively optimize the NeRF with sketch-conditioned guidance.
These results show that our method achieves state-of-the-art performance in terms of sketch similarity and text alignment.

\subsection{Ablation Study}
\label{Ablation Study}


\noindent\textbf{Variation of Sketches.}
Fig.~\ref{fig:tesser} shows the results of the proposed method with different sketches as input.
Note that all the prompts are fixed to \textit{"a teapot"}, our method can produce high-fidelity 3D objects resembling the specific sketches (e.g., the handle shape, the spout position, and the body size).
This demonstrates that our method can generate diverse 3D objects specified by varied sketches.

\begin{figure}[t]
    \centering
    \begin{subfigure}{\linewidth}
        \includegraphics[width=\linewidth]{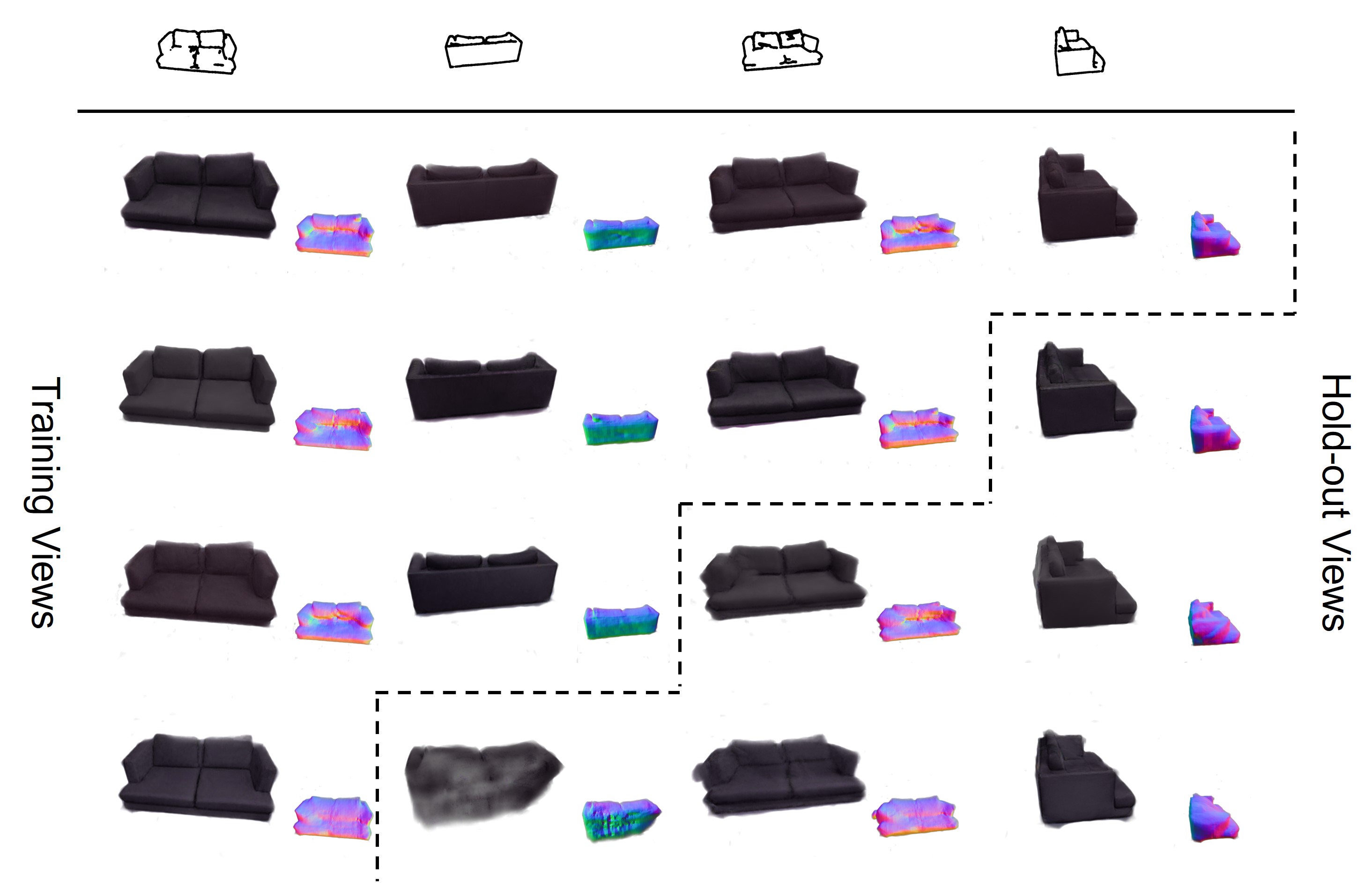}
    \end{subfigure}
    \caption{
        Generated objects using various quantities of sketch images.
        From \textit{top} to \textit{bottom}, each row shows the results (of rendered RGB and normal images) using 24, 12, 6, and 3 sketch images as input, respectively.
    }
    \label{fig-results-number-of-sketch}
    \vspace{-0.4cm}
\end{figure}

\noindent\textbf{Quantity of Sketch Images.}
We generated a 3D sofa by using sketches with quantities of 24, 12, 6, and 3, respectively.
Fig.~\ref{fig-results-number-of-sketch} shows that our method can produce 3D consistent objects with plausible geometry even using only 3 sketches.
This is because the random viewpoint regularization ensures the reality of the generated object.


\begin{figure}[hbpt]
    \centering
    \begin{subfigure}{0.24\linewidth}
        \begin{minipage}[t]{\linewidth}
            \includegraphics[width=\linewidth]{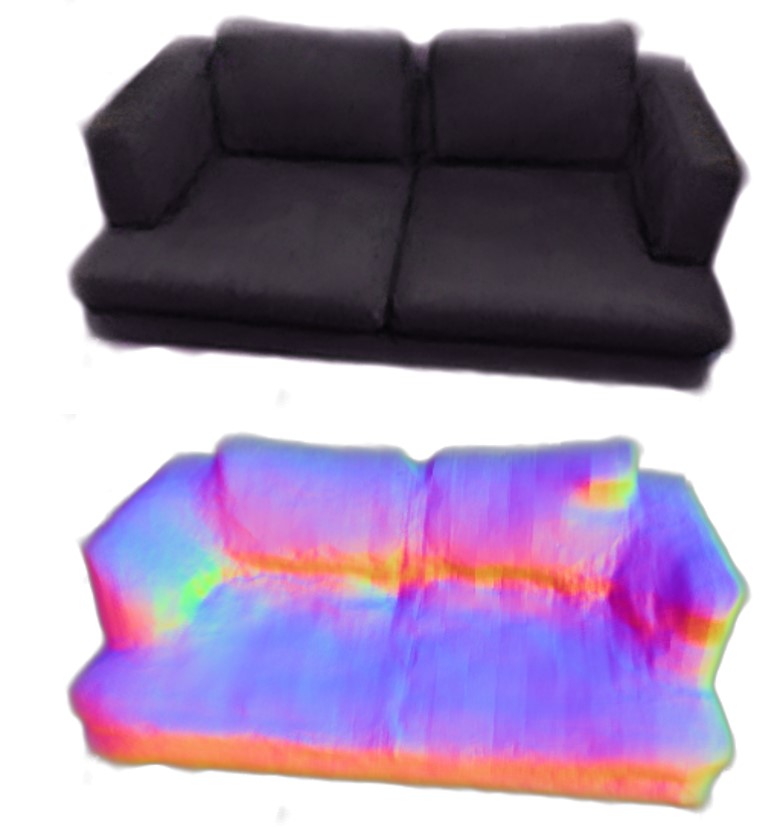}
        \end{minipage}
        \caption*{0}
    \end{subfigure}
    \begin{subfigure}{0.24\linewidth}
        \begin{minipage}[t]{\linewidth}
            \includegraphics[width=\linewidth]{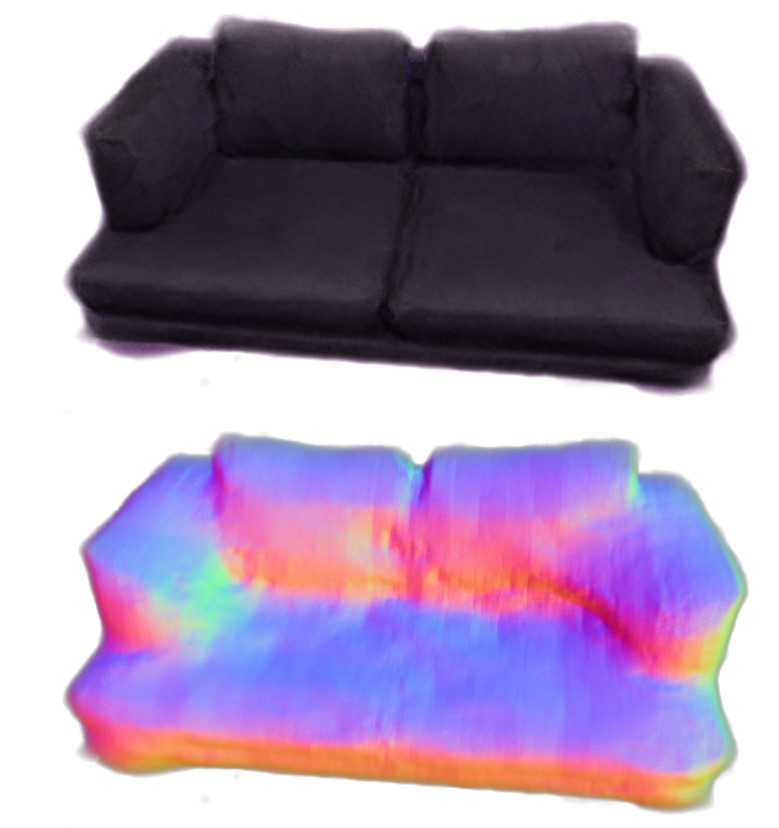}
        \end{minipage}
        \caption*{0.01}
    \end{subfigure}
    \begin{subfigure}{0.24\linewidth}
        \begin{minipage}[t]{\linewidth}
            \includegraphics[width=\linewidth]{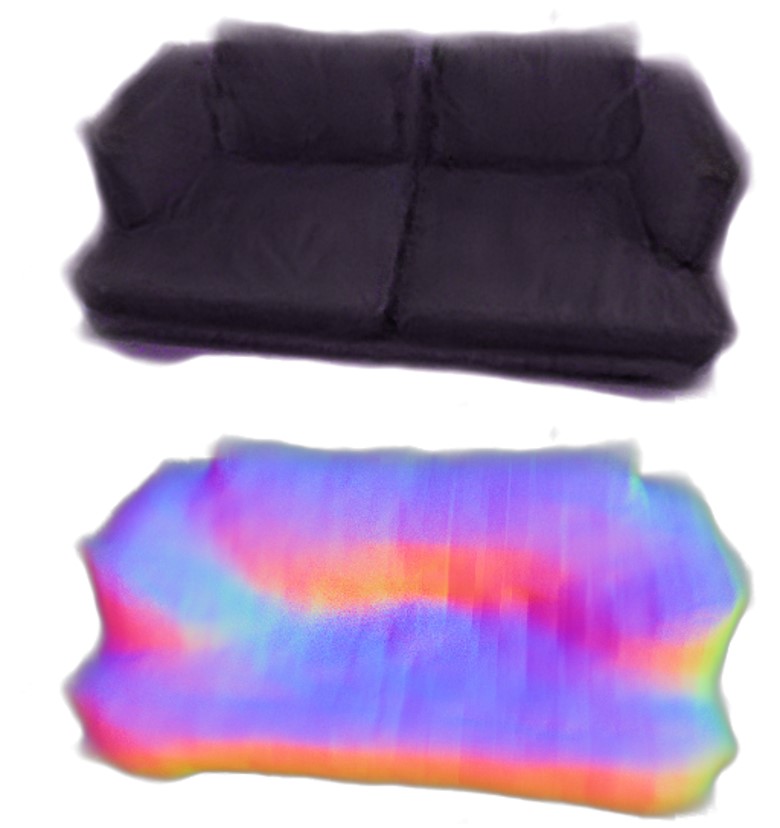}
        \end{minipage}
        \caption*{0.02}
    \end{subfigure}
    \begin{subfigure}{0.24\linewidth}
        \begin{minipage}[t]{\linewidth}
            \includegraphics[width=\linewidth]{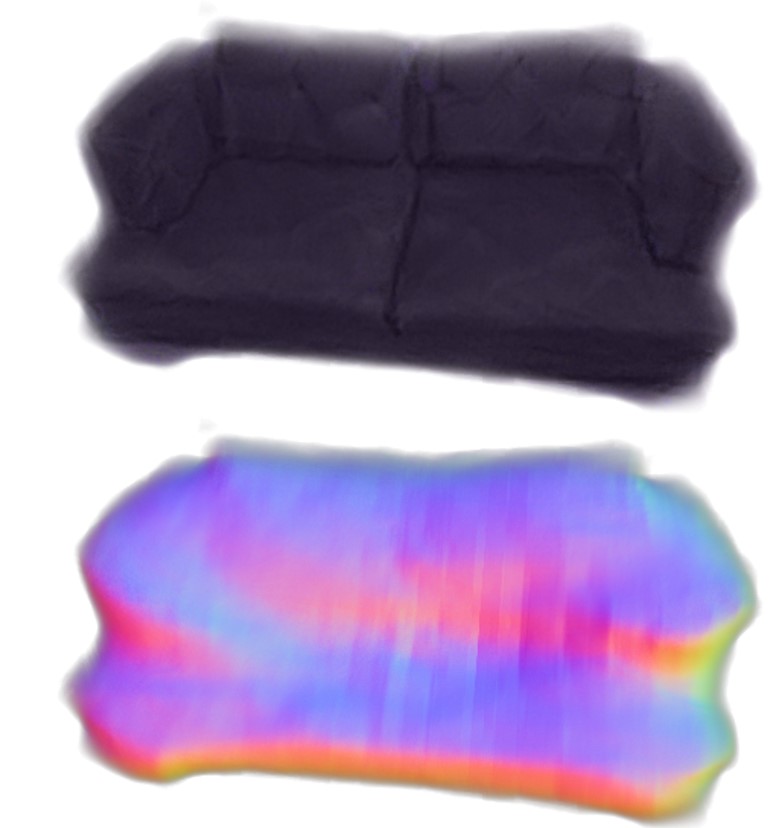}
        \end{minipage}
        \caption*{0.03}
    \end{subfigure}
    
    \caption{
        Rendered RGB and normal images of objects generated under various intensities noises of sketch poses.
    }
    \label{fig-results-noisy-pose}
    \vspace{-0.4cm}
\end{figure}

\noindent\textbf{Noisy Poses.}
We conducted experiments to study the robustness of our method for the noise poses.
We perturb the sketch poses with different intensities (i.e., 0, 0.01, 0.02, and 0.03), such that the multi-view sketches cannot maintain consistency.
Fig.~\ref{fig-results-noisy-pose} shows that higher-intensity noises lead to blurry contents and coarse geometry, because inconsistent multi-view sketches introduce higher uncertainty in 3D generation.

\section{Conclusion}
In this paper, we proposed a novel multi-view sketch-guided text-to-3D generation method (namely \textit{Sketch2NeRF}) to generate high-fidelity 3D content that resembles given sketches.
In particular, we proposed a novel synchronous generation and reconstruction optimization strategy to effectively fit the NeRF from 2D ControlNet guidance.
In addition, we employ the annealed time schedule to further improve the quality of NeRF in the generation procedure.
We collected two types of datasets to evaluate the proposed methods.
Experimental results demonstrate the superior performance of our method on fine-grained control provided by sketches. 
{
    \small
    \bibliographystyle{ieeenat_fullname}
    \bibliography{main}
}


\end{document}